\documentclass[runningheads]{llncs}

\usepackage{eccv}

\usepackage{eccvabbrv}

\usepackage{graphicx}
\usepackage{booktabs}
\usepackage{multirow}
\usepackage{xcolor}
\usepackage{wrapfig}
\usepackage{float}

\usepackage[accsupp]{axessibility}  

\usepackage{hyperref}

\usepackage{orcidlink}

\begin{document}

\definecolor{firebrick}{RGB}{178, 34, 34}
\definecolor{goldenrod}{RGB}{218, 165, 32}
\definecolor{limegreen}{RGB}{50, 205, 50}


\title{PYRA: Parallel Yielding Re-Activation for Training-Inference Efficient Task Adaptation} 

\titlerunning{PYRA: Parallel Yielding Re-Activation}

\author{Yizhe Xiong\inst{1,2}\orcidlink{0009-0001-5233-9466} \and
Hui Chen\inst{2}\thanks{Corresponding Author}\orcidlink{0000-0003-4180-5801} \and
Tianxiang Hao\inst{1,2}\orcidlink{0000-0002-1952-6083} \and
Zijia Lin\inst{1}\orcidlink{0000-0002-1390-7424} \and
Jungong Han\inst{2,3}\orcidlink{0000-0003-4361-956X} \and
Yuesong Zhang\inst{4}\orcidlink{0009-0008-8991-9609} \and
Guoxin Wang\inst{4}\orcidlink{0000-0002-6193-0864} \and
Yongjun Bao\inst{4}\orcidlink{0000-0002-7816-0587} \and
Guiguang Ding\inst{1,2}\orcidlink{0000-0003-0137-9975}}

\authorrunning{Y.~Xiong et al.}

\institute{School of Software, Tsinghua University, Beijing, China \and
BNRist, Tsinghua University, Beijing, China \and
Department of Automation, Tsinghua University, Beijing, China \and
JD.com, Beijing, China\\
\email{\{xiongyizhe2001,beyondhtx,jungonghan77\}@gmail.com}\\
\email{huichen@mail.tsinghua.edu.cn}\\
\email{\{zhangyuesong1,wangguoxin14,baoyongjun\}@jd.com}\\
\email{linzijia07@tsinghua.org.cn\quad dinggg@tsinghua.edu.cn}}

\maketitle

\begin{abstract}
  Recently, the scale of transformers has grown rapidly, which introduces considerable challenges in terms of training overhead and inference efficiency in the scope of task adaptation. 
  Existing works, namely Parameter-Efficient Fine-Tuning (PEFT) and model compression, have separately investigated the challenges. 
  However, PEFT cannot guarantee the inference efficiency of the original backbone, especially for large-scale models. Model compression requires significant training costs for structure searching and re-training. 
  Consequently, a simple combination of them cannot guarantee accomplishing both training efficiency and inference efficiency with minimal costs. In this paper,  we propose a novel Parallel Yielding Re-Activation (PYRA) method for such a challenge of training-inference efficient task adaptation. PYRA first utilizes parallel yielding adaptive weights to comprehensively perceive the data distribution in downstream tasks. A re-activation strategy for token modulation is then applied for tokens to be merged, leading to calibrated token features. Extensive experiments demonstrate that PYRA outperforms all competing methods under both low compression rate and high compression rate, demonstrating its effectiveness and superiority in maintaining both training efficiency and inference efficiency for large-scale foundation models. Our code is available at \url{https://github.com/THU-MIG/PYRA}.
\end{abstract}

\section{Introduction}
\label{sec:1_introduction}

Vision transformers~\cite{dosovitskiy2020image} have made a profound impact across various domains of computer vision, such as image classification~\cite{dosovitskiy2020image,chen2021crossvit,wang2021pyramid,zhai2022scaling,ding2023exploring,tian2024clip}, object detection~\cite{carion2020end,zhu2020deformable,liu2021swin,song2023graphalign,song2023graphalign++}, image segmentation~\cite{strudel2021segmenter,zheng2021rethinking,cheng2022masked,wang2023repvit,zhou2022context}, etc. 
In recent years, the scale of vision transformers has grown to be billion-parameters~\cite{zhai2022scaling,kirillov2023segment,xiong2024temporal}. Consequently, adapting such models with vast scales into downstream tasks presents increasingly complex challenges, particularly in real-world deployment scenarios. Two critical concerns have been widely acknowledged as primary obstacles in implementing large-scale transformers for downstream applications~\cite{houlsby2019parameter,hao2023consolidator,jiang2023llmlingua,song2024ba}: (1) the training overhead when fine-tuning on downstream tasks, and (2) the inference efficiency after model deployment.

\begin{figure}[t]
  \centering
   \includegraphics[width=1.0\linewidth]{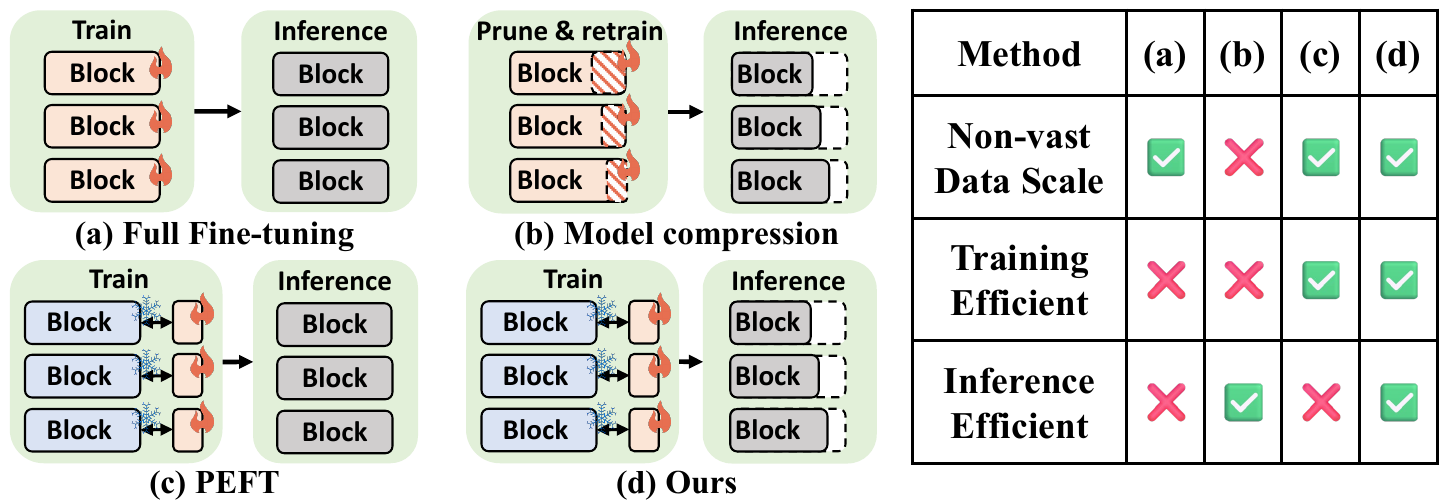}
   \caption{(a) Full fine-tuning trains all parameters on downstream tasks and utilizes the trained model for inference, thereby lacking efficiency in both training and inference stages. (b) Model compression employs pruning to enhance inference efficiency, but the pruned model necessitates extensive re-training on large-scale data. (c) PEFT freezes the model backbone and only fine-tunes a small amount of parameters, yet retains the inference complexity. (d) Our training-inference efficient task adaptation incorporates the advantages of all existing pipelines by training inference-efficient models with minimal tunable parameters.}
   \label{fig:teaser}
\end{figure}

Specifically, first, conventional fine-tuning methods, which necessitate adjusting all parameters of the model (\ie, Full Fine-tuning), suffer from unaffordable consumption of GPU resources and training time given the extensive scales of the foundation models~\cite{dosovitskiy2020image,zhai2022scaling,kirillov2023segment}. Researchers have delved into Parameter-Efficient Fine-Tuning (PEFT)~\cite{houlsby2019parameter,jia2022visual,lian2022scaling,hu2021lora,hao2023consolidator,hao2023re} algorithms, which generally freeze the pre-trained models and only tune extra small parameters, leading to great reduction of training time and storage overhead. The second issue pertains to inference efficiency, requiring the deployed model to promptly process the input data. The computational complexity of the models significantly influences achieving satisfactory inference throughput. Representative solutions for this matter encompass model compression methods, including model pruning~\cite{chen2021chasing,yu2022unified,wang2022vtc,chen2024image}, knowledge distillation~\cite{hinton2015distilling,ahn2019variational,tung2019similarity,wei2024video,wei2024more}, model quantization~\cite{courbariaux2015binaryconnect,lin2017towards,cai2020zeroq}, etc.

In the literature, these two intriguing topics are investigated separately. PEFT methods either identify a subset of tunable parameters in the backbone for fine-tuning~\cite{zaken2021bitfit} or introduce learnable parameters to the frozen backbone during fine-tuning~\cite{houlsby2019parameter,jia2022visual,lian2022scaling,hu2021lora,hao2023consolidator}. While effectively reducing training costs, most of these methods inevitably escalate computational complexity, resulting in inefficient inference. Model compression methods frequently require significant computational resources to identify optimal structures for pruning. After pruning, a comprehensive re-training process using a substantial amount of data is crucial to prevent significant performance degradation. Therefore, model compression methods are typically inefficient in training efficiency.

\begin{wrapfigure}{R}{0.6\textwidth}
   \includegraphics[width=0.6\textwidth, trim={1.5cm 0 2.2cm 1.75cm}, clip]{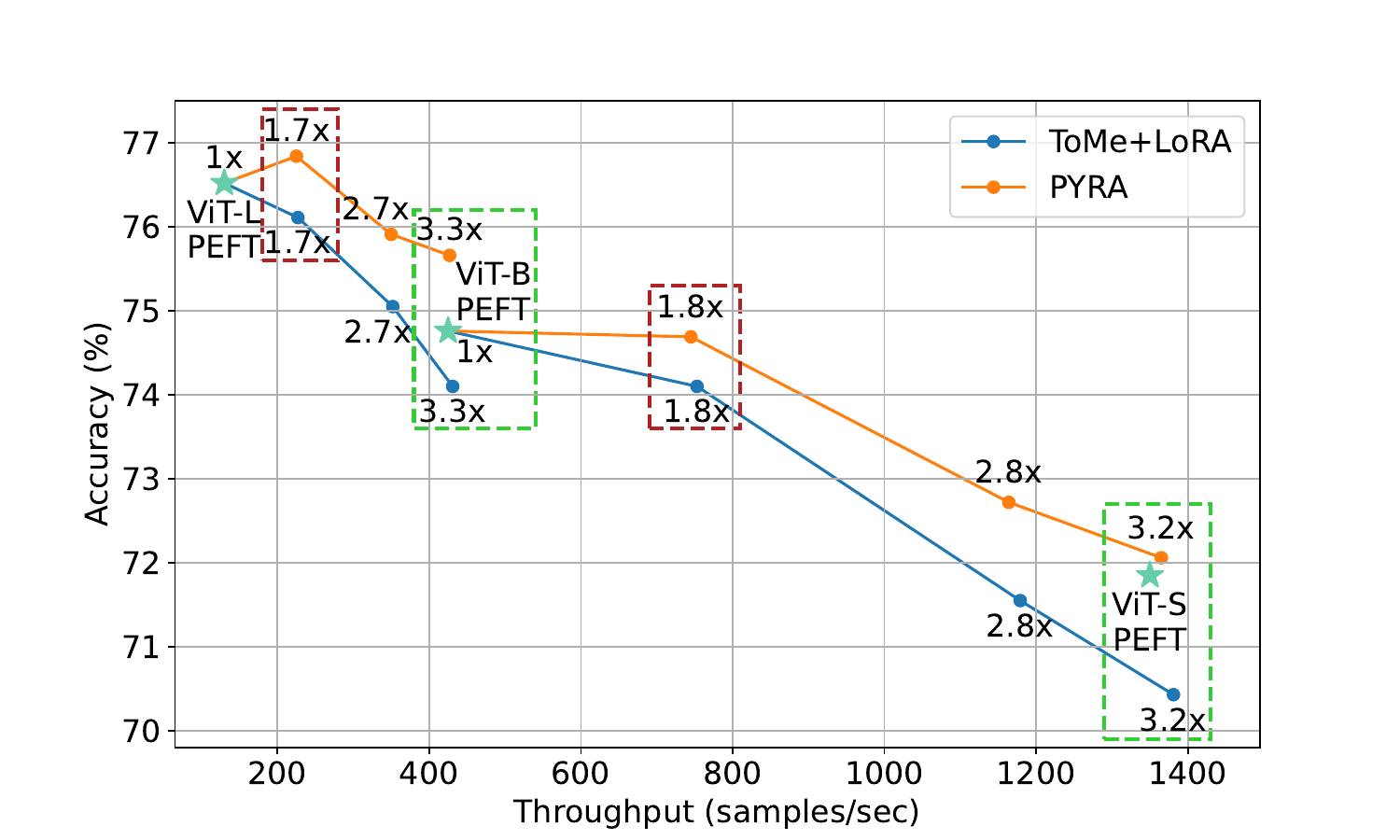}
   \caption{Comparisons between simply combining ToMe~\cite{bolya2022token} and LoRA~\cite{hu2021lora} and our proposed PYRA. \textcolor{firebrick}{\textbf{Red boxes}} represent the performance drop problem in low compression rates. \textcolor{limegreen}{\textbf{Green boxes}} represent the adverse compression in high compression rates. See \cref{sec:4_experiment} for more results.}
   \label{fig:combine}
\end{wrapfigure}

These observations naturally lead to the question: \textit{can we achieve both training efficiency and inference efficiency simultaneously for downstream tasks?} 
We refer to this challenge as \textbf{Training-Inference Efficient Task Adaptation} (seeing \cref{fig:teaser}). 
Exploring this issue can enable us to conveniently deploy the advanced large-scale foundation models in real-world downstream applications \texttt{with minimal costs}, which is appealing and essential for the widespread implementation of foundation models.
A straightforward solution is combining PEFT and model compression. However, for efficient task adaptation, a heavy re-training stage is unaffordable. Consequently, simply combining PEFT and model compression can easily suffer from substantial performance drops. 
For instance, we could integrate LoRA~\cite{hu2021lora}, a notable PEFT method that introduces a small low-rank adapter, with ToMe~\cite{bolya2022token}, a parameter-free model compression technique for vision transformers\footnote{In \cref{sec:4_experiment}, we show that ToMe+LoRA is a neat and strong solution.}. 
As shown in \cref{fig:combine}, under lower compression rates (around 1.7$\times$), performance on both backbones (ViT-L/16 and ViT-B/16) present slight drops (<1\%) compared to directly conducting PEFT on the backbone, indicating that the ToMe+LoRA combination serves as a basic solution within lower compression rate range, yet performance improvements are also demanded. Under high compression rates (>3.0$\times$), the performance quickly drops and is even inferior to directly fine-tuning the small-scale backbone with corresponding throughput. We term this phenomenon as \textit{Adverse Compression}. Both phenomena indicate that directly combining existing works cannot effectively address the new challenges.

In this paper, we propose a novel \textbf{P}arallel \textbf{Y}ielding \textbf{R}e-\textbf{A}ctivation method (\textbf{PYRA}) designed for training-inference efficient task adaptation using vision transformers. Generally, the proposed PYRA follows the token merging paradigm~\cite{marin2021token,bolya2022token,chen2023diffrate} for inference efficiency. For effective task adaptation, we propose to modulate token features from both feature tokens and feature channels and parallelly yield weights for tokens to be merged by small modulation weight generators.  
These parallel yielding weights can comprehensively perceive the data distribution in downstream tasks. They are then applied to features through re-activation, resulting in adaptive token modulation. Thanks to such a token modulation strategy, PYRA can adaptively calibrate the learned feature distribution for downstream tasks with low computational complexity, ultimately leading to effective training and efficient inference. 

We conduct extensive experiments to verify the effectiveness of our PYRA. We show that, under a low compression rate with $\sim$1.7$\times$ speedup, PYRA introduces negligible performance drops. Under a high compression rate with $>$3.0$\times$ speedup, PYRA eliminates the adverse compression gap. We emphasize that considering the scenario of high compression rates is practical, given the substantial size of current transformers. Concurrently, the specific model sizes required by downstream applications may not match any publicly available models, thus requiring the acquisition through high compression rates while keeping the performance comparable. In this regard, our approach represents an effective method for obtaining small models in the absence of pre-trained parameters for smaller-scale models.

Overall, we summarize our contribution as follows.
\begin{itemize}
    \item We propose a novel challenge termed training-inference efficient task adaptation, in which the inference efficiency of large-scale transformers is escalated during parameter-efficient task adaptation.
    \item We propose PYRA for training-inference efficient task adaptation, which enhances the perception of feature distribution via token modulation. We generate parallel yielding decoupled weights to comprehensively perceive the feature distributions. We apply a re-activation strategy to modulate tokens to be merged for calibrated token features.   
    \item Extensive experiments show that our PYRA outperforms all competing methods under both low compression rate and high compression rate. Further analysis shows that PYRA is effective across a series of different transformer backbones and model scales, well demonstrating the effectiveness and superiority of our method.
\end{itemize}

\section{Related Works}
\label{sec:2_related}

\noindent\textbf{Parameter-Efficient Fine-Tuning (PEFT) for Task Adaptation.}
Transferring large-scale transformers to downstream tasks has been a popular topic in 
computer vision~\cite{jia2022visual,jie2022convolutional,lian2022scaling,zhang2022new,zhang2022neural,chen2022adaptformer,hao2023consolidator,xiong2023confidence}. PEFT methods either locate a subset of parameters inside the model for fine-tuning~\cite{zaken2021bitfit}, or inject human designed modules to the original model structure. 
Specifically, adapters~\cite{houlsby2019parameter,pfeiffer2020adapterfusion,jie2022convolutional,lyu2023one} are a type of MLP module with a bottleneck in the middle. 
Prompt-tuning~\cite{lester2021power,li2021prefix,liu2021p,ding2022delta,jia2022visual,hao2024qprompt} insert learnable tokens to model input to generate task-specific outputs.
SSF~\cite{lian2022scaling} tunes additional scaling and shifting parameters. 
LoRA~\cite{hu2021lora}, AdaptFormer~\cite{chen2022adaptformer}, and Consolidator~\cite{hao2023consolidator} add lightweight modules as bypasses. These modules can be merged with the original backbone for no extra inference cost.
In the scope of training-inference efficient task adaptation, PEFT methods retain model inference cost, resulting in extreme difficulty for model deployment. To solve the problem, we conduct adaptive token merging for PEFT. While achieving promising adaptation performance under both low and high compression rates, our method inherits the advantages of PEFT methods.

\noindent\textbf{Model Compression.}
Model compression is often applied on large-scale models to acquire smaller-scale models with comparable performance. Mainstream approaches of model compression include model pruning~\cite{chen2021chasing,yu2022unified,wang2022vtc,bolya2022token,chen2023diffrate,wang2023cait,zhou2024ppr}, knowledge distillation~\cite{hinton2015distilling,ahn2019variational,tung2019similarity} and model quantization~\cite{courbariaux2015binaryconnect,lin2015neural,lin2017towards,cai2020zeroq}. 
For transformer models, model pruning methods can be roughly grouped into two categories: channel pruning and token pruning. Channel pruning methods~\cite{chen2021chasing,zhu2021vision,chavan2022vision,yu2022unified} reduce the number of parameters, channels, heads, or blocks. 
Recently, token pruning has emerged as another mainstream approach. Several works have attempted to prune tokens for vision transformers (ViTs).
Among these methods, token pooling~\cite{marin2021token} uses a slow k-means approach that does not work for an off-the-shelf model. ToMe~\cite{bolya2022token} constructs bipartite graphs and merges token pairs with the most weighted connections. DiffRate~\cite{chen2023diffrate} combines token pruning and token merging with searched optimal token reduction rates for each layer. 
Although achieving promising results, most existing model compression methods fail when combining with PEFT for training-inference efficient task adaptation since they usually involve a heavy training stage. Model pruning methods demand full re-training after pruning to restore performance. Knowledge distillation necessitates training a small-scale model from scratch, demanding large amount of data. As for model quantization, although no heavy training is demanded, post-training quantization~\cite{liu2021post,yao2022zeroquant} achieves poor performance compared to quantization-aware training~\cite{li2022q,huang2023variation} that demands full re-training on the quantized model. 
As a comparison, our proposed PYRA achieves promising performance on compressed models under the restrictions of PEFT, surpassing the baselines of simply combining model compression and PEFT.

\section{Methodology}

\subsection{Preliminaries}
\label{sec:3_1_preliminaries}
\noindent\textbf{ViT Model.} 
In this paper, we mainly focus on the training-inference efficient task adaptation of ViT models~\cite{dosovitskiy2020image,touvron2021training,he2022masked}. A ViT model consists of $L$ identical encoder blocks, each of which consists of a multi-head self-attention (MHSA) module and a feed-forward network (FFN). Formally, an input image $\mathbf{x}$ is reshaped and linear projected to $N$ tokens $\mathbf{x}=[t_1,t_2,\cdots,t_N]$ in $D$ dimensions. For simplicity, we omit the classification token ([CLS]) and the distillation token~\cite{touvron2021training}. 
For encoder block $l$, we denote the input as $\mathbf{x}^l=[t^l_1,t^l_2,\cdots,t^l_{N^{l-1}}]\in\mathbb{R}^{N^{l-1}\times D}$ and the output as $\hat{\mathbf{x}}^l=[\hat{t}^l_1,\hat{t}^l_2,\cdots,\hat{t}^l_{N^l}]\in\mathbb{R}^{N^{l}\times D}$.
For MHSA, the input tokens are first processed by three FC layers to generate $Q$,$K$, and $V$ matrices, and the output is calculated by $\text{Softmax}(\frac{QK^T}{\sqrt{D}})V$ before being projected by another FC layer.  
For FFN, the tokens are projected by two FC layers.
Our method mainly focuses on the input tokens $\mathbf{x}^l$ before feeding them to the MHSA module.

\noindent\textbf{LoRA.}
LoRA~\cite{hu2021lora} is a widely employed PEFT method for task adaptation. Vision transformers consist of large dense parameter matrices. When adapting to a specific task, the updates to the matrices are in small subspaces and can be modeled with low-rank decompositions. LoRA trains only the decomposed matrices during fine-tuning. Specifically, for dense matrix $W_0 \in \mathbb{R}^{d\times k}$ and input $x$, the modified forward pass of updated $W_0$ is:
\begin{equation}
    \hat{x}=W_0x+BAx,
    \label{eq:lora}
\end{equation}
where $B\in \mathbb{R}^{d\times h}$ and $A\in \mathbb{R}^{h\times k}$. During inference, $BA$ can be merged with $W_0$ for no extra computation overhead.

\noindent\textbf{Token Merging.} 
Token merging~\cite{bolya2022token,chen2023diffrate} is a parameter-free compression technique for vision transformers. It is orthogonal to the transformer structure and capable of flexibly changing the compression rate. 
Specifically, the input tokens $\mathbf{x}^l$ of the $l$-th ViT block are randomly separated before the MHSA:
\begin{equation}
    G^l_1=[t^l_{i_1},t^l_{i_2},\cdots,t^l_{i_s}],\quad G^l_2=[t^l_{j_1},t^l_{j_2},\cdots,t^l_{j_s}],\quad 2s=N^{l-1}.
    \label{eq:baseline_grouping}
\end{equation}
Then, for each $t^l_{i_\bullet}$ token, the most similar $t^l_{j_\bullet}$ token is matched to it via cosine similarity. From $G^l_1$, $r$ tokens with the most similar connections to tokens in $G^l_2$ are selected to form token pairs $(t^l_{m_k},t^l_{n_k})$, where $t^l_{m_k}\in G^l_1$, $t^l_{n_k}\in G^l_2$, and $k=1,2,\cdots,r$. Note that $\{m_k\}$ is a re-indexed subset of $\{i_\bullet\}$. Formally,
\begin{equation}
    \{t_{m_k}^l\} = \arg \underset{\hat{t}^l_i}{\text{Top}r}(\underset{j}{\max}\,\frac{\hat{t}^l_i\cdot t^l_j}{\|\hat{t}^l_i\|\cdot\|t^l_j\|}),\quad
    t_{n_k}^l = \arg \underset{t_j^l}{\max}(\frac{t^l_{m_k}\cdot t^l_j}{\|t^l_{m_k}\|\cdot\|t^l_j\|})
    \label{eq:baseline_matching}
\end{equation}
Previous works~\cite{marin2021token,bolya2022token,chen2023diffrate}, generally merge tokens via average pooling, cutting down the computational cost by decreasing the token number by $r$ in layer $l$.

We employ the above techniques as our baseline method, in which we attach token merging while fine-tuning LoRA for task adaptation. 
These methods are selected due to their advantages being highly compatible with training-inference efficient task adaptation. First, token merging is parameter-free and training-free, and does not change model structures, which retains the storage-efficient advantage of PEFT. Besides, we choose LoRA for its popularity, simplicity, and mergeability (not introducing extra FLOPs during inference). Experimental results show that ToMe+LoRA is a strong baseline (see \cref{sec:4_experiment}).

\subsection{PYRA: Parallel Yielding Re-Activation}
\label{sec:3_2_pyra} 

Conventional fine-tuning methods excel at guiding the model to dynamically align with the target data distribution in downstream tasks by extensively adjusting parameters. However, in the context of training-inference efficient task adaptation, only a fraction of parameters can be fine-tuned, posing significant challenges in accurately capturing the nuances of data distribution. While reducing model complexity through token merging shows promise in enhancing inference efficiency, it introduces the risk of information loss during layer-wise processing in vision transformers. This loss is difficult to rectify due to the constrained understanding of data distribution in the efficient task adaptation scenario. Therefore, a straightforward combination of token merging with PEFT algorithms for achieving training-inference efficient task adaptation may not yield optimal results, as depicted in \cref{fig:combine}. 

Here, we propose \textbf{P}arallel \textbf{Y}ielding \textbf{R}e-\textbf{A}ctivation (\textbf{PYRA}) to adaptively modulate token features to enhance the perception of data distribution during token merging. Specifically, inside PYRA, the weights for adaptive merging are first yielded in a parallel manner through a pair of lightweight learnable vectors in each ViT block. These generated weights are then applied to modulate tokens to be merged through re-activation. 
As a result, PYRA enables adaptive calibration of the learned feature distribution with low computational complexity.

\begin{figure*}[t]
  \centering
  \includegraphics[width=1.0\linewidth]{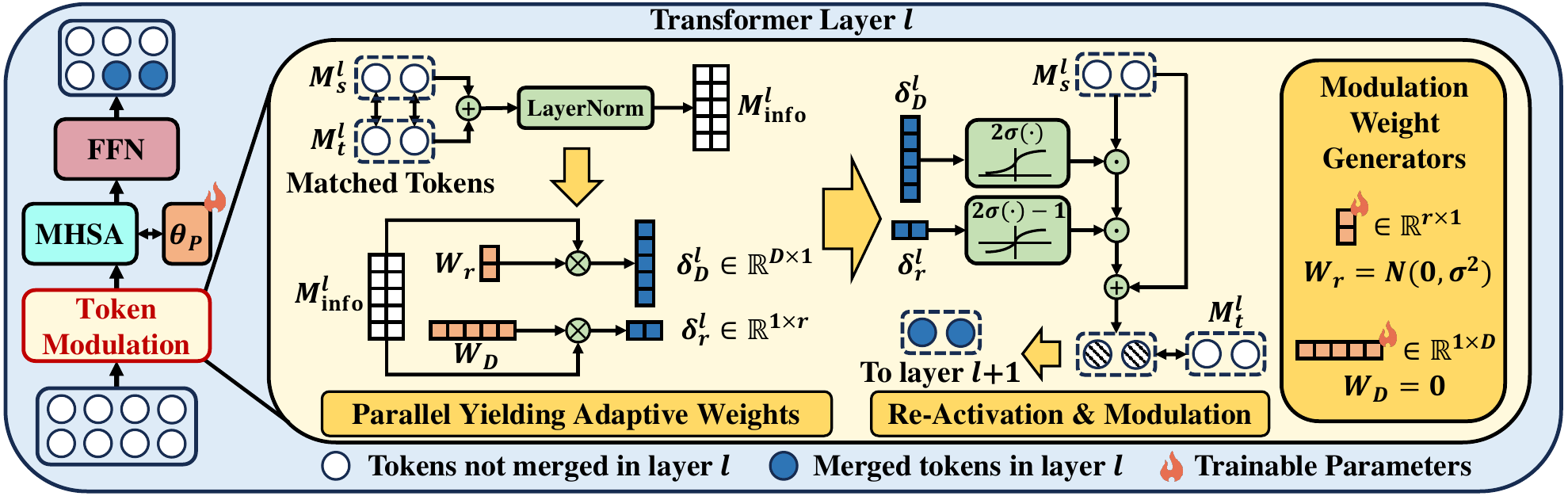}
  \caption{The pipeline of our PYRA. PYRA conducts token modulation before the MHSA module in each transformer block. Inside PYRA, a pair of learnable modulation weight generators are leveraged to generate adaptive modulation weights parallelly. After that, generated weights modulate tokens through re-activation. The generators in PYRA can be trained along with the LoRA module $\theta_p$ in an end-to-end manner.}
  \label{fig:method}
\end{figure*}

\textbf{Parallel Yielding Adaptive Weights.}
We aim to optimize the merging process of each chosen token pair. 
Inspired by feature modulation~\cite{wu2018group,perez2018film}, we propose to modulate token features before merging. 
Formally, for encoder block $l$ with $r$ pairs of tokens in $D$ dimensions to be merged, we group the $t^l_{m_k}$ and $t^l_{n_k}$ tokens as token matrices $M^l_s=[t^l_{m_1}, \cdots, t^l_{m_r}]$ and $M^l_t=[t^l_{n_1}, \cdots, t^l_{n_r}]$, where $M^l_s,M^l_t \in \mathbb{R}^{D\times r}$. We learn a modulation matrix $W^l\in \mathbb{R}^{D\times r}$ that adaptively modulates tokens at the granularity of each channel. 
We emphasize that directly learning $W^l$ is redundant~\cite{hu2021lora} and cannot adaptively satisfy the conditions with different images and token pairs. Therefore, we further specify the goal as learning decoupled weights $\delta^l_D=f_D(M^l_s,M^l_t)\in\mathbb{R}^{D\times 1}$ for feature channels and $\delta^l_r=f_r(M^l_s,M^l_t)\in\mathbb{R}^{1\times r}$ for feature tokens, where $W^l=\delta^l_D\delta^l_r$, separately. 
We propose to generate $\delta^l_D$ and $\delta^l_r$ in a parallel yielding manner.

Specifically, for layer $l$ with $r$ tokens to be merged, we create two learnable vectors as the modulation weight generator inside the transformer block: $W^l_r\in \mathbb{R}^{r\times 1}$ and $W^l_D\in \mathbb{R}^{1\times D}$. To guarantee that $W^l_r$ and $W^l_D$ digest the token features from both tokens in a token pair, we first normalize the sum of token pairs to construct the token information matrix:
\begin{equation}
      M^l_\text{info} = \text{LayerNorm}(M^l_s+M^l_t) \in \mathbb{R}^{D\times r}.
  \label{eq:token_info}
\end{equation}
We normalize the distribution of $M^l_\text{info}$ tokens by leveraging the $\text{LayerNorm}(\cdot)$ operation to enable smoother gradients when training $W^l_r$ and $W^l_D$. With the token information matrix, we then yield the adaptive weights $\delta^l_D$ and $\delta^l_r$ parallelly:
\begin{equation}
    \begin{aligned}
      \delta^l_D & = M^l_\text{info}W^l_r \in \mathbb{R}^{D\times 1} \\
      \delta^l_r & = W^l_DM^l_\text{info} \in \mathbb{R}^{1\times r}.
    \end{aligned}
  \label{eq:pyra_m}
\end{equation}

\textbf{Re-Activation for Token Modulation.}
Simply generating $\delta^l_D$ and $\delta^l_r$ via matrix multiplication still faces several possible issues. First, 
no measures have taken to ensure that $\delta^l_D$ and $\delta^l_r$ stay in a normal range. Second, decoupling weights to feature tokens and feature channels results in a low-rank modulation weight matrix $W^l$, which exhibits limited expressive capacity and thus could be unable to optimally modulate tokens in complicated data distributions. To cope with these issues, we conduct token modulation in a re-activation strategy. Specifically, we first broadcast $\delta^l_D$ to $\hat{\delta}^l_D\in \mathbb{R}^{D\times r}$ and conduct sigmoid activation $\sigma(\cdot)$ on it, and then modulate $M^l_s$ for an intermediate modulation result:
\begin{equation}
      \hat{M}^l_s = 2\sigma(\hat{\delta}_D^l)\odot M^l_s.
  \label{eq:re_activation_1}
\end{equation}
where $\odot$ is Hadamard product. 
$\hat{M}^l_s$ is then modulated with sigmoid-activated broadcast weight $\hat{\delta}_r^l\in \mathbb{R}^{D\times r}$ again to acquire the modulated tokens:
\begin{equation}
      M^l_s \leftarrow M^l_s+(2\sigma(\hat{\delta}_r^l)-1)\odot \hat{M}^l_s,
  \label{eq:re_activation_2}
\end{equation}
Note that we use the original tokens $M^l_s$ to create a residual connection that preserves the gradient flow during training. 
The modulated $M^l_s$ tokens are then merged with $M^l_t$ with average pooling. We use a random Gaussian initialization for generator $W_r^l$ and zero for generator $W_D^l$, so re-activation is equivalent to identity transformation at the beginning of training. 
Token modulation is conducted only on $M_s^l$ to guarantee the parallelism in training and inference, as $t^l_{m_k}$ tokens are distinct while different $t^l_{n_k}$ might point to the same $t^l_j$ token.

\textbf{Discussion.} 
Our PYRA effectively enhances the calibration of features for downstream tasks. Specifically, by utilizing parallel yielding adaptive weights, PYRA effectively decouples the modulation weight from the feature token and the feature channel, \ie, $\delta_{D}$ and $\delta_{r}$, enabling comprehensive perception of the feature distribution in downstream tasks. 
Furthermore, considering the challenge of capturing an accurate feature distribution for parameter-efficient training, such smooth re-activations by \cref{eq:re_activation_1} and \cref{eq:re_activation_2} can constrain the negative impact on weight values brought by 
limited perception of feature distributions, thus leading to better token modulation for improving feature representations in downstream tasks.
As a result, the proposed PYRA can maintain discriminative information to an utmost degree during token merging, leading to improved performance while achieving complexity reduction.

\subsection{Complexity Analysis}
\label{sec:3_3_analysis}

We present the parameter and computation complexity for introducing PYRA to PEFT for task adaptation. In each ViT block, PYRA introduces $D+r$ training parameters and $4rD$ extra FLOPs. For a ViT model with $L$ layers and $R$ total merged tokens, in total, our PYRA introduces $LD+R$ extra training parameters and conducts $4RD$ extra FLOPs beside PEFT.

\begin{table}[t]
\fontsize{7.}{8.7}
\selectfont
\setlength{\tabcolsep}{0.8mm}
\centering
\makeatletter\def\@captype{table}\makeatother\caption{The complexity comparisons between conducting PEFT with and without PYRA. The FLOPs metric is obtained during inference.}
\label{tab:complexity_analysis}
\begin{tabular}{c|c|c|cc|cc}
\toprule
Model & Metric & Total & PEFT w/o PYRA & (\%) & PEFT w. PYRA & (\%) \\
\midrule
\multirow{2}{*}{ViT-Base} & \# params & 86M & 0.29M & 0.34\% & 0.30M & 0.35\% \\
 & FLOPs & 16.37G & 16.37G & 100\% & 8.15G & 49.79\% \\
\midrule
\multirow{2}{*}{ViT-Large} & \# params & 303M & 1.18M & 0.39\% & 1.20M & 0.40\% \\
 & FLOPs & 57.37G & 57.37G & 100\% & 28.76G & 50.13\% \\
\bottomrule
\end{tabular}
\end{table}

To better demonstrate the training-inference efficiency of our PYRA, we compare the complexity between attaching PYRA to task adaptation via PEFT (here we employ LoRA~\cite{hu2021lora}) and plainly conducting PEFT without PYRA. As shown in \cref{tab:complexity_analysis}, PYRA keeps the training efficient feature of PEFT by introducing only a tiny amount of training parameters, which however, results in a substantial efficiency boost for inference, \ie, at around 50\% for both ViT-B and ViT-L models. Results in \cref{sec:4_experiment} show that PYRA achieves comparable performance to the PEFT counterpart without PYRA. This well indicates that PYRA is an effective method for training-inference efficient task adaptation.

\section{Experiments}
\label{sec:4_experiment}  

\subsection{Experimental Setting}
\label{sec:4_1_exp_set}

\textbf{Datasets.}
We conduct extensive experiments on task adaptation benchmarks to verify the effectiveness of PYRA within the challenge of training-inference efficient task adaptation. Specifically, we choose the VTAB-1k~\cite{zhai2019large} benchmark for the evaluation. VTAB-1k is a challenging benchmark that consists of 19 different tasks from diverse domains: 1) natural images captured in the actual world; 2) specialized images from professional fields; and 3) structured synthesized images. Each task only contains 800 training samples and 200 validation samples. 

\noindent\textbf{Models.}
We choose two ViT backbones pre-trained on the ImageNet-21K~\cite{dosovitskiy2020image}, \ie, ViT-L/16 and ViT-B/16, for comparison. Additionally, we also generalize our PYRA to different backbones and pre-train methods, including DeiT-B~\cite{touvron2021training} and ViT backbones pre-trained by MAE~\cite{he2022masked}. 

\noindent\textbf{Implementation Details.}
We choose LoRA~\cite{hu2021lora} as the PEFT module for all methods due to its simplicity and mergeability.
\textbf{For ease of explanation, we omit the ``+LoRA'' when comparing different methods.} We append LoRA only on the $Q$,$K$, and $V$ projection matrices, and apply the training schedule for LoRA following~\cite{zhang2022neural,hao2023consolidator}. The generators are trained along with LoRA modules. During inference, we merge the LoRA module to the backbone. All throughputs are measured during inference on a GeForce RTX 3090 GPU. More details can be found in the supplementary materials.

\begin{table*}[t]
\fontsize{1.0}{6.0} 
\selectfont
\setlength{\tabcolsep}{0.3mm}
\caption{Results on the VTAB-1k~\cite{zhai2019large} benchmark under low compression rate. \textbf{Bold} and \underline{underline} denote the best and second-best accuracy within compression methods.}
\label{tab:vtab_results_low}
\hspace{-0.44cm}
\begin{tabular}{ccc|ccccccc|cccc|cccccccc|c}
\toprule
& & & \multicolumn{7}{c}{\textbf{Natural}} \vline & \multicolumn{4}{c}{\textbf{Specialized}} \vline & \multicolumn{8}{c}{\textbf{Structured}} \vline & \\
\rotatebox{90}{Method} & \rotatebox{90}{\# params} & \rotatebox{90}{Throughput}  & \rotatebox{90}{Cifar100} & \rotatebox{90}{Caltech101} & \rotatebox{90}{DTD}  & \rotatebox{90}{Flowers102} & \rotatebox{90}{Pets} & \rotatebox{90}{SVHN} & \rotatebox{90}{Sun397} & \rotatebox{90}{Camelyon} & \rotatebox{90}{EuroSAT} & \rotatebox{90}{Resisc45} & \rotatebox{90}{Retinopathy} & \rotatebox{90}{Clevr-Count} & \rotatebox{90}{Clevr-Dist} & \rotatebox{90}{DMLAB} & \rotatebox{90}{KITTI-Dist} & \rotatebox{90}{dSpr-Loc} & \rotatebox{90}{dSpr-Ori} & \rotatebox{90}{sNORB-Azim} & \rotatebox{90}{sNORB-Ele} & \rotatebox{90}{Average} \\
\midrule
\multicolumn{23}{c}{\textbf{Model: ViT-B/16 (Throughput: 425)}} \\
\midrule
PEFT & 0.34\% & 425 & 67.1 & 90.2 & 69.4 & 99.1 & 90.5 & 85.7 & 54.1 & 83.1 & 95.8 & 84.3 & 74.6 & 82.2 & 69.2 & 50.1 & 79.2 & 81.8 & 47.1 & 31.1 & 42.6 & 74.76 \\
\hline
RaP & 3.43\% & 654 & 25.9 & 68.4 & 53.3 & 64.0 & 57.4 & 71.3 & 21.5 & 75.8 & 87.9 & 59.3 & 73.6 & 43.1 & 53.8 & 26.3 & 60.5 & 73.5 & 25.5 & 16.7 & 27.9 & 55.57 \\
SPViT & 4.46\% & 567 & 41.6 & 75.4 & 61.1 & 83.2 & 66.2 & 56.1 & 28.3 & 79.3 & 94.2 & 73.3 & 73.6 & 70.6 & 61.5 & 42.4 & 67.8 & 75.4 & \textbf{50.5} & 28.9 & 31.3 & 64.16 \\
DiffRate & 0.35\% & 709 & 37.1 & 84.6 & 63.7 & 96.7 & 86.2 & 32.6 & 48.2 & 78.9 & 85.8 & 67.0 & 73.7 & 32.9 & 29.8 & 34.1 & 55.7 & 12.6 & 16.0 & 13.1 & 21.5 & 55.82 \\
ToMe & 0.34\% & 753 & \underline{64.6} & \textbf{90.4} & \underline{67.9} & \underline{98.5} & \underline{89.8} & \underline{83.9} & \textbf{53.2} & \underline{82.6} & \underline{94.7} & \textbf{83.5} & \underline{74.9} & \underline{81.9} & \textbf{69.8} & \underline{49.2} & \underline{76.9} & \textbf{81.9} & \underline{46.5} & \underline{31.0} & \textbf{43.1} & \underline{74.10} \\
PYRA & 0.35\% & 745 & \textbf{67.5} & \underline{90.3} & \textbf{69.3} & \textbf{98.9} & \textbf{90.0} & \textbf{84.6} & \underline{53.1} & \textbf{83.3} & \textbf{95.7} & \underline{83.3} & \textbf{75.2} & \textbf{82.6} & \underline{68.9} & \textbf{50.8} & \textbf{80.0} & \underline{81.8} & 45.8 & \textbf{32.2} & \underline{42.8} & \textbf{74.69} \\
\midrule
\multicolumn{23}{c}{\textbf{Model: ViT-L/16 (Throughput: 130)}} \\
\midrule
PEFT & 0.39\% & 130 & 77.1 & 91.4 & 73.4 & 99.5 & 91.3 & 89.6 & 57.6 & 85.9 & 96.1 & 87.3 & 76.1 & 83.1 & 63.0 & 50.7 & 82.1 & 81.7 & 53.5 & 32.2 & 36.6 & 76.52 \\
\hline
RaP & 1.95\% & 196 & 43.2 & 87.9 & 62.6 & 52.8 & 81.7 & 86.7 & 34.7 & 78.4 & 92.4 & 73.3 & 73.6 & 68.0 & 59.6 & 46.9 & \underline{82.4} & 75.5 & 43.6 & 24.5 & 25.7 & 65.64 \\
SPViT & 2.47\% & 188 & 48.1 & 87.5 & 65.2 & 94.4 & 77.4 & 80.9 & 38.8 & 79.9 & 93.9 & 79.8 & 74.3 & 78.2 & \textbf{65.8} & 47.4 & 74.1 & \underline{82.3} & 50.3 & 31.0 & 37.9 & 70.22 \\
DiffRate & 0.39\% & 221 & 50.9 & 86.8 & 70.3 & 97.8 & 88.3 & 39.0 & 52.3 & 80.2 & 87.2 & 72.2 & 74.2 & 32.6 & 32.3 & 36.5 & 57.4 & 22.8 & 26.6 & 15.2 & 23.4 & 59.53 \\
ToMe & 0.39\% & 227 & \underline{76.1} & \underline{91.1} & \underline{72.3} & \underline{99.2} & \textbf{91.7} & \underline{89.2} & \underline{56.4} & \underline{86.4} & \underline{95.1} & \underline{86.6} & \underline{75.1} & \underline{82.4} & 61.9 & \underline{50.9} & 81.4 & 81.6 & \textbf{53.5} & \underline{33.4} & \underline{36.8} & \underline{76.11} \\
PYRA & 0.40\% & 225 & \textbf{76.6} & \textbf{91.3} & \textbf{73.2} & \textbf{99.3} & \underline{91.5} & \textbf{89.4} & \textbf{57.1} & \textbf{86.9} & \textbf{95.9} & \textbf{87.1} & \textbf{76.2} & \textbf{83.2} & \underline{63.2} & \textbf{52.8} & \textbf{83.1} & \textbf{82.5} & \underline{52.6} & \textbf{34.8} & \textbf{39.0} & \textbf{76.84} \\
\bottomrule
\end{tabular}
\end{table*}

\begin{table*}[t]
\fontsize{1.0}{6.0} 
\selectfont
\setlength{\tabcolsep}{0.3mm}
\caption{Results on the VTAB-1k~\cite{zhai2019large} benchmark under high compression rate. \textbf{Bold} and \underline{underline} denote the best and second-best accuracy within compression methods. $*$: As a comparison of similar throughputs, we compare ViT-B/16 with PEFT on ViT-S/16, and ViT-L/16 with PEFT on ViT-B/16.}
\label{tab:vtab_results_high}
\hspace{-0.5cm}
\begin{tabular}{ccc|ccccccc|cccc|cccccccc|c}
\toprule
& & & \multicolumn{7}{c}{\textbf{Natural}} \vline & \multicolumn{4}{c}{\textbf{Specialized}} \vline & \multicolumn{8}{c}{\textbf{Structured}} \vline & \\
\rotatebox{90}{Method} & \rotatebox{90}{\# params} & \rotatebox{90}{Throughput}  & \rotatebox{90}{Cifar100} & \rotatebox{90}{Caltech101} & \rotatebox{90}{DTD}  & \rotatebox{90}{Flowers102} & \rotatebox{90}{Pets} & \rotatebox{90}{SVHN} & \rotatebox{90}{Sun397} & \rotatebox{90}{Camelyon} & \rotatebox{90}{EuroSAT} & \rotatebox{90}{Resisc45} & \rotatebox{90}{Retinopathy} & \rotatebox{90}{Clevr-Count} & \rotatebox{90}{Clevr-Dist} & \rotatebox{90}{DMLAB} & \rotatebox{90}{KITTI-Dist} & \rotatebox{90}{dSpr-Loc} & \rotatebox{90}{dSpr-Ori} & \rotatebox{90}{sNORB-Azim} & \rotatebox{90}{sNORB-Ele} & \rotatebox{90}{Average} \\
\midrule
\multicolumn{23}{c}{\textbf{Model: ViT-B/16 (Throughput: 425)}} \\
\midrule
PEFT$^*$ & 0.34\% & 1350 & 57.7 & 88.2 & 70.1 & 98.7 & 88.7 & 85.7 & 44.9 & 81.4 & 94.7 & 84.6 & 73.6 & 81.6 & 64.1 & 48.1 & 80.0 & 72.9 & 38.4 & 22.9 & 37.7 & 71.85 \\
\hline
RaP & 0.86\% & 1029 & 24.3 & 40.1 & 34.5 & 41.8 & 40.5 & 21.7 & 11.4 & 75.8 & 86.5 & 35.1 & 73.8 & 49.6 & 49.7 & 28.1 & 39.4 & 13.8 & 15.4 & 12.4 & 26.9 & 42.60 \\
SPViT & 4.46\% & 944 & 23.7 & 67.9 & 51.9 & 69.9 & 53.2 & 19.6 & 13.1 & 71.9 & 81.3 & 67.9 & \underline{74.7} & 53.5 & 61.9 & 39.5 & 57.4 & 45.0 & 34.5 & 11.1 & 23.2 & 52.49 \\
DiffRate & 0.35\% & 1308 & 23.2 & 73.0 & 55.7 & 87.9 & 66.7 & 27.2 & 29.3 & 78.1 & 77.8 & 53.1 & 73.6 & 29.7 & 28.6 & 31.7 & 52.6 & 11.5 & 17.2 & 11.3 & 20.3 & 49.29 \\
ToMe & 0.34\% & 1381 & \textbf{54.2} & \underline{87.8} & \underline{65.5} & \underline{96.1} & \underline{81.7} & \underline{79.7} & \textbf{45.2} & \underline{79.4} & \underline{93.6} & \underline{76.3} & 73.8 & \underline{78.3} & \underline{65.7} & \underline{48.0} & \underline{71.3} & \underline{80.0} & \underline{45.8} & \underline{30.9} & \underline{41.2} & \underline{70.43} \\
PYRA & 0.35\% & 1365 & \underline{54.0} & \textbf{89.3} & \textbf{67.1} & \textbf{96.5} & \textbf{84.0} & \textbf{81.8} & \underline{44.6} & \textbf{81.2} & \textbf{94.6} & \textbf{79.5} & \textbf{75.1} & \textbf{79.9} & \textbf{67.0} & \textbf{49.2} & \textbf{76.9} & \textbf{82.6} & \textbf{47.8} & \textbf{31.9} & \textbf{42.0} & \textbf{72.06} \\
\midrule
\multicolumn{23}{c}{\textbf{Model: ViT-L/16 (Throughput: 130)}} \\
\midrule
PEFT$^*$ & 0.34\% & 425 & 67.1 & 90.2 & 69.4 & 99.1 & 90.5 & 85.7 & 54.1 & 83.1 & 95.8 & 84.3 & 74.6 & 82.2 & 69.2 & 50.1 & 79.2 & 81.8 & 47.1 & 31.1 & 42.6 & 74.76\\
\hline
RaP & 0.65\% & 301 & 17.7 & 37.1 & 27.0 & 46.2 & 33.3 & 23.2 & 13.3 & 76.5 & 74.2 & 54.4 & 73.6 & 50.4 & 31.4 & 25.7 & 49.8 & 53.1 & 25.5 & 13.4 & 26.0 & 44.11 \\
SPViT & 2.47\% & 289 & 54.0 & 87.6 & 65.5 & 94.8 & 74.9 & 32.6 & 38.6 & 81.8 & \textbf{95.3} & 78.0 & 74.0 & 72.8 & \underline{61.2} & 46.9 & 70.2 & 77.1 & 47.4 & 31.3 & 28.6 & 66.90 \\
DiffRate & 0.39\% & 416 & 47.4 & 73.5 & 54.1 & 84.3 & 60.2 & 19.6 & 22.2 & 50.0 & 64.6 & 42.8 & 18.2 & 31.5 & 31.9 & 31.1 & 37.3 & 22.0 & 17.7 & 14.8 & 21.4 & 40.48 \\
ToMe & 0.39\% & 431 & \underline{71.0} & \underline{90.9} & \underline{70.4} & \underline{98.3} & \underline{88.5} & \underline{87.2} & \textbf{52.4} & \underline{82.9} & \underline{94.5} & \underline{83.1} & \underline{75.0} & \underline{80.7} & 61.1 & \underline{48.9} & \underline{76.9} & \underline{80.8} & \underline{53.0} & \underline{32.1} & \underline{35.2} & \underline{74.10} \\
PYRA & 0.40\% & 427 & \textbf{71.6} & \textbf{91.8} & \textbf{71.1} & \textbf{98.5} & \textbf{89.7} & \textbf{88.1} & \underline{52.2} & \textbf{85.1} & \textbf{95.3} & \textbf{84.6} & \textbf{75.7} & \textbf{80.9} & \textbf{63.0} & \textbf{51.7} & \textbf{82.0} & \textbf{82.0} & \textbf{54.2} & \textbf{36.0} & \textbf{41.2} & \textbf{75.66} \\
\bottomrule
\end{tabular}
\end{table*}

\begin{table}[t]
\begin{minipage}[t]{0.47\linewidth}
\fontsize{7.2}{8.6}
\selectfont
\setlength{\tabcolsep}{0.6mm}
\begin{center}
\caption{VTAB-1k~\cite{zhai2019large} results on both compression rates for ViT-L (MAE) (Throughput: 130).}
\label{tab:vtab_mae}
\begin{tabular}{l|ccc|c}
\toprule
Model& \rotatebox{90}{Method} & \rotatebox{90}{\# params} & \rotatebox{90}{Throughput} & \rotatebox{90}{Average} \\
\midrule
\parbox{1.72cm}{ViT-L (MAE)} & PEFT & 0.39\% & \parbox{0.55cm}{130} & 75.96 \\
\hline
\parbox{1.72cm}{\multirow{4}{*}{ViT-L (MAE)}} & RaP~\cite{DBLP:conf/cvpr/LiA0GTG22} & 2.01\% & \parbox{0.55cm}{182} & 67.55 \\
& DiffRate~\cite{chen2023diffrate} & 0.39\% & \parbox{0.55cm}{221} & 46.91 \\
& ToMe~\cite{bolya2022token} & 0.39\% & \parbox{0.55cm}{227} & \underline{74.97} \\
& PYRA & 0.40\% & \parbox{0.55cm}{225} & \textbf{76.13} \\
\midrule
\midrule
\parbox{1.72cm}{ViT-B (MAE)} & PEFT & 0.34\% & \parbox{0.55cm}{425} & 70.23 \\
\hline
\parbox{1.72cm}{\multirow{4}{*}{ViT-L (MAE)}} & RaP~\cite{DBLP:conf/cvpr/LiA0GTG22} & 0.76\% & \parbox{0.55cm}{298} & 52.91 \\
& DiffRate~\cite{chen2023diffrate} & 0.39\% & \parbox{0.55cm}{416} & 48.03 \\
& ToMe~\cite{bolya2022token} & 0.39\% & \parbox{0.55cm}{431} & \underline{68.20} \\
& PYRA & 0.40\% & \parbox{0.55cm}{427} & \textbf{70.33} \\
\bottomrule
\end{tabular}
\end{center}
\end{minipage} \qquad
\begin{minipage}[t]{0.47\linewidth}
\fontsize{7.2}{8.6}
\selectfont
\setlength{\tabcolsep}{0.6mm}
\begin{center}
\caption{VTAB-1k~\cite{zhai2019large} results on both compression rates for DeiT-B (Throughput: 431).}
\label{tab:vtab_deit}
\begin{tabular}{l|ccc|c}
\toprule
Model& \rotatebox{90}{Method} & \rotatebox{90}{\# params} & \rotatebox{90}{Throughput} & \rotatebox{90}{Average} \\
\midrule
\parbox{1.72cm}{DeiT-B} & PEFT & 0.34\% & \parbox{0.55cm}{431} & 73.76 \\
\hline
\parbox{1.72cm}{\multirow{4}{*}{DeiT-B}} & RaP~\cite{DBLP:conf/cvpr/LiA0GTG22} & 3.57\% & \parbox{0.55cm}{695} & 62.34 \\
& DiffRate~\cite{chen2023diffrate} & 0.35\% & \parbox{0.55cm}{734} & 52.10 \\
& ToMe~\cite{bolya2022token} & 0.34\% & \parbox{0.55cm}{747} & \underline{72.83} \\
& PYRA & 0.35\% & \parbox{0.55cm}{740} & \textbf{73.55} \\
\midrule
\midrule
\parbox{1.72cm}{DeiT-S} & PEFT & 0.34\% & \parbox{0.55cm}{1332} & 70.01 \\
\hline
\parbox{1.72cm}{\multirow{4}{*}{DeiT-B}} & RaP~\cite{DBLP:conf/cvpr/LiA0GTG22} & 1.09\% & \parbox{0.55cm}{1187} & 57.70 \\
& DiffRate~\cite{chen2023diffrate} & 0.35\% & \parbox{0.55cm}{1314} & 44.51 \\
& ToMe~\cite{bolya2022token} & 0.34\% & \parbox{0.55cm}{1351} & \underline{68.68} \\
& PYRA & 0.35\% & \parbox{0.55cm}{1341} & \textbf{70.13} \\
\bottomrule
\end{tabular}
\end{center}
\end{minipage}
\end{table}

\subsection{Performance comparison on Task Adaptation Benchmark}
\label{sec:4_2_benchmark}

We verify the effectiveness of PYRA for training-inference efficient task adaptation on (1) low compression rate: comparing the performance of different methods under the same sparsity ratio (here we set sparsity ratio=50\%); (2) high compression rate: comparing the performance between the competing methods and the smaller-scale model with similar throughput levels. We leverage ViT-L/16 and ViT-B/16 for these inspections.
The competing methods include RaP~\cite{DBLP:conf/cvpr/LiA0GTG22}, SPViT~\cite{DBLP:conf/eccv/KongDMMNSSYRTQW22}, DiffRate~\cite{chen2023diffrate}, and ToMe~\cite{bolya2022token}. Comparisons to more baselines are in the supplementary materials. For RaP and SPViT, we first train the backbone with LoRA on downstream tasks, then we prune the models, and lastly we re-train the parameters attached by the pruning method and LoRA. DiffRate and ToMe can be employed with LoRA during fine-tuning, while DiffRate demands ImageNet-21K for searching the optimal compression schedule.

\textbf{PYRA on low compression rate.}
Results on low compression rate are reported in \cref{tab:vtab_results_low}. Overall, while achieving one of the best speedups on throughput, our PYRA is the best performed method compared to other competing methods using the lowest level of training parameters. Counting results on both backbones, our PYRA achieves the best or second-best performance on 37 of 38 dataset metrics. Compared to directly conducting PEFT on the backbone, while all competing methods cause worse results, our PYRA achieves comparable adaptation performance on ViT-B/16, and even outperforms ViT-L/16. 
The above results convincingly demonstrate that our PYRA successfully sets a new benchmark, \ie reaching comparable performance as the uncompressed model, on low compression rate for training-inference efficient task adaptation while accelerating the model to 1.75$\times$ speedup with only 0.4\% training parameters.

\begin{figure}[t]
  \centering
  \includegraphics[width=1.0\linewidth, trim={1cm 0 0 0}, clip]{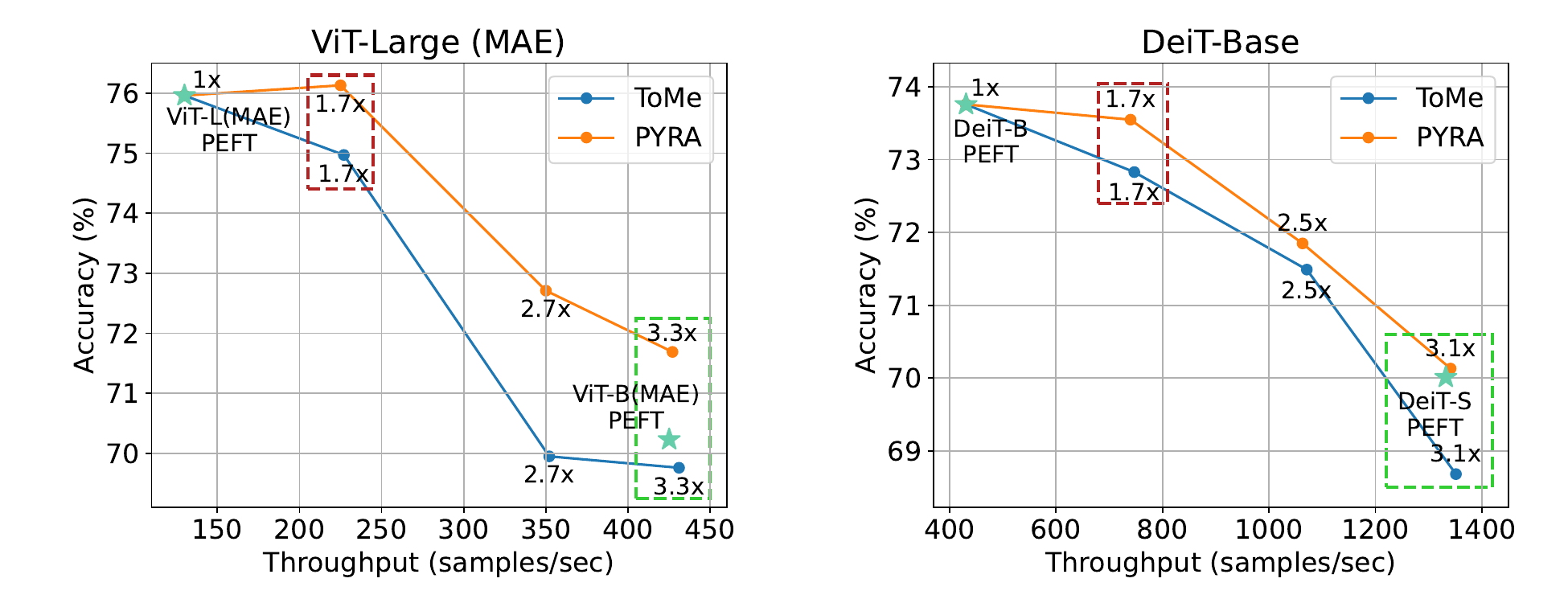}
  \caption{Comparisons between PYRA and ToMe~\cite{bolya2022token} under different compression rates for ViT-Large (MAE) and DeiT-Base. \textcolor{firebrick}{\textbf{Red boxes}}: PYRA mitigates the performance drops under low compression rate. \textcolor{limegreen}{\textbf{Green boxes}}: PYRA elinimates adverse compression under high compression rate.}
  \label{fig:curve_other_exps}
\end{figure}

\textbf{PYRA on high compression rate.}
Results are reported in \cref{tab:vtab_results_high}. With comparable throughputs to the smaller-scale model and minimal training parameters, PYRA outperforms all competing methods and achieves \underline{the best} performance on 35 of 38 dataset metrics. Compared to the smaller-scale model with comparable throughput, PYRA successfully outperforms it. On the compressed ViT-L/16, PYRA even surpasses ViT-B/16 by 0.9\%. This shows that PYRA, as the state-of-the-art, effectively bridges the adverse compression gap between compressed large-scale models and small-scale models, as mentioned in \cref{sec:1_introduction}. Therefore, PYRA is an applicable alternative for acquiring a smaller-scale model (3.2$\times$ speedup) on downstream tasks through efficient training (0.4\% training parameters) when no pre-trained small-scale model is available.

\textbf{PYRA on self-supervised ViT backbone.}
We conduct the experiments on both compression rates on self-supervised ViT-L/16 (MAE)~\cite{he2022masked}. As shown in \cref{tab:vtab_mae}, PYRA significantly surpasses all competing methods. Under low compression rate, PYRA outperforms directly transferring the uncompressed backbone. Under high compression rate, PYRA also eliminates adverse compression. These results show that PYRA generalizes well for self-supervised visual models.

\textbf{PYRA on different architectures.}
To further verify the generalizability of PYRA, we conduct the above experiments on the DeiT-B~\cite{touvron2021training}. As shown in \cref{tab:vtab_deit}, PYRA achieves comparable performance to the uncompressed model under low compression rate and eliminates adverse compression under high compression rate, indicating its generalizability to other transformer architectures.

\begin{table}[t]
\fontsize{7.}{8.7}
\selectfont
\setlength{\tabcolsep}{0.8mm}
\centering
\makeatletter\def\@captype{table}\makeatother\caption{Ablation study results for PYRA on ViT-B/16 under high compression rate. Here for \# params we report only the parameters introduced by token modulation.}
\label{tab:ablation_study}
\begin{tabular}{c|ccc|c|ccc|c}
\toprule
Method & $W_r$ & $W_D$ & Activation & \# params & Natural & Specialized & Structured & Average\\
\midrule
Baseline & $\times$ & $\times$ & $\times$ & 0 & 72.87 & 80.78 & 57.64 & 70.43 \\
Plain $W_r$ & $\surd$ & $\times$ & $\times$ & 0.19K & 72.90 & 81.07 & 57.66 & 70.54 \\
$\sigma(\cdot)$ \& $W_r$ & $\surd$ & $\times$ & $\surd$ & 0.19K & 73.18 & 81.69 & 57.65 & 70.84 \\
Plain $W_D$ & $\times$ & $\surd$ & $\times$ & 8.45K & 73.09 & 81.13 & 58.43 & 70.88 \\
$\sigma(\cdot)$ \& $W_D$ & $\times$ & $\surd$ & $\surd$ & 8.45K & 73.31 & \underline{82.17} & 58.44 & 71.31 \\
Plain $W_r$ \& $W_D$ & $\surd$ & $\surd$ & $\times$ & 8.64K & \underline{73.77} & 81.37 & \underline{58.81} & \underline{71.32} \\
PYRA & $\surd$ & $\surd$ & $\surd$ & 8.64K & \textbf{73.91} & \textbf{82.60} & \textbf{59.66} & \textbf{72.06} \\
\bottomrule
\end{tabular}
\end{table}

\textbf{PYRA consistently yields better models of different throughputs.}
To further show the superiority of PYRA for different compression rates, we compress the chosen backbones for a series of speedups and compare the results with the strongest baseline, ToMe~\cite{bolya2022token}, under similar values of throughputs. As shown in \cref{fig:combine,fig:curve_other_exps}, PYRA consistently outperforms ToMe and flattens the accuracy-throughput curve. Therefore, our PYRA is applicable to acquire task-specific smaller-scale models of different throughputs consistently in the absence of pre-trained parameters for smaller-scale models.

\subsection{Ablation Studies}
\label{sec:4_3_ablation}
We do controlled experiments to identify the effect of individual components in PYRA, \ie, the generators $W_r$ and $W_D$, and the sigmoid activation $\sigma(\cdot)$ applied on the generated modulation weights. 
Specifically, when using only a single generator, we omit \cref{eq:re_activation_1} and replace $\hat{\delta}_r^l$ in \cref{eq:re_activation_2} with the broadcast weight generated by the employed generator. When using the generators without sigmoid activation, we simply remove the $\sigma$ in \cref{eq:re_activation_1} and \cref{eq:re_activation_2} while keeping other calculations intact.
We choose the ImageNet-21K pre-trained ViT-B/16 to carry out the ablation studies under high compression rates as in \cref{tab:vtab_results_high}. 
Results are shown in \cref{tab:ablation_study}. Here the baseline method refers to conducting token merging~\cite{bolya2022token,chen2023diffrate} while fine-tuning LoRA~\cite{hu2021lora} for task adaptation.
First, both $W_r$ and $W_D$, no matter whether sigmoid activation is attached, lead to performance gains, and employing them simultaneously outperforms using them individually. This proves the superiority of our parallel yielding strategy. Second, compared to the corresponding counterpart without activations, employing sigmoid activation always leads to significant improvements, indicating the effectiveness of the re-activation. Overall, our PYRA yields the best adaptation performance, showing that our strategies are effective and complementary.

\begin{table}[t]
\begin{minipage}{0.49\linewidth}
\fontsize{7.}{8.7}
\selectfont
\setlength{\tabcolsep}{0.8mm}
\centering
\makeatletter\def\@captype{table}\makeatother\caption{Adaptation performance comparison to $W_{r\times D}$ unadaptable to tokens. We report token modulation parameters.}
\label{tab:analysis_adaptive}
\resizebox{\textwidth}{!}{
\begin{tabular}{cc|ccc|c}
\toprule
\rotatebox{90}{Method} & \rotatebox{90}{\# params} & \rotatebox{90}{Natural} & \rotatebox{90}{Specialized} & \rotatebox{90}{Structured} & \rotatebox{90}{Average}\\
\midrule
Baseline & 0 & 72.9 & 80.8 & 57.6 & 70.43 \\
$W^l_{D\times r}$ & 70.66K & \underline{73.8} & \underline{81.5} & \underline{59.2} & \underline{71.49} \\
PYRA & 8.64K & \textbf{73.9} & \textbf{82.6} & \textbf{59.7} & \textbf{72.06} \\
\bottomrule
\end{tabular}}
\end{minipage}\hfill
\begin{minipage} {0.49\linewidth}
\fontsize{7.}{8.7}
\selectfont
\setlength{\tabcolsep}{0.8mm}
\centering
\makeatletter\def\@captype{table}\makeatother\caption{Adaptation performance comparison to the common gated generator. We report token modulation parameters.}
\label{tab:analysis_gate}
\resizebox{\textwidth}{!}{
\begin{tabular}{cc|ccc|c}
\toprule
\rotatebox{90}{Method} & \rotatebox{90}{\# params} & \rotatebox{90}{Natural} & \rotatebox{90}{Specialized} & \rotatebox{90}{Structured} & \rotatebox{90}{Average}\\
\midrule
Baseline & 0 & \underline{72.9} & 80.8 & 57.6 & 70.43 \\
Gated & 73.73K & 72.8 & \underline{81.5} & \underline{58.9} & \underline{71.06} \\
PYRA & 8.64K & \textbf{73.9} & \textbf{82.6} & \textbf{59.7} & \textbf{72.06} \\
\bottomrule
\end{tabular}}
\end{minipage}\hfill
\end{table}

\subsection{Further Analysis on Different Designs}
\label{sec:4_4_analysis}

We conduct further analysis on ViT-B/16 model pre-trained on ImageNet-21K.
The baseline here refers to training LoRA with token merging~\cite{bolya2022token,chen2023diffrate} attached.
More analysis experiments can be found in the supplementary materials.

\textbf{Impact of Making Modulation Weights Adaptive.}
In our PYRA, we train modulation weight generators $W^l_D$ and $W^l_r$ to conduct adaptive modulation on different merging tokens. To prove the necessity of making modulation weights adaptive, we compare PYRA with the approach of directly training final modulation weights $W^l_{D\times r}$ for each layer, and conduct token modulation as $M^l_s \leftarrow M^l_s+(2\sigma(W^l_{D\times r})-1)\odot M^l_s$.
For fair comparisons, we inherit the sigmoid activation and the residual connection. As shown in \cref{tab:analysis_adaptive}, although with more training parameters, training $W_{D\times r}$ is still inferior to our PYRA. This indicates that PYRA is a more effective strategy to conduct adaptive token modulation.

\textbf{Compare with the Common Gated Generator.}
We compare PYRA with setting the commonly-applied gated-style trainable module~\cite{HochSchm97,chung2014empirical,kong2021spvit,xu2023parameter} as modulation weight generator. Formally, for each ViT block, we insert a learnable two-layer MLP module ($\text{MLP}^l(\cdot)$) with input dimension $D$, hidden dimension $d\ll D$ to ensure parameter-efficient training and fast inference, and output dimension $D$. To generate modulation weights, we feed the information matrix $M^l_{\text{info}}$ into the MLP, and modulate the merging tokens thereafter: $\delta^l = \text{MLP}(M^l_\text{info}) \in \mathbb{R}^{D\times r}, M^l_s \leftarrow M^l_s+(2\sigma(\delta^l-1)\odot M^l_s.$
We set $d=4$ for all layers. As shown in \cref{tab:analysis_gate}, both the gated generator and our PYRA achieve performance gains, while our PYRA surpasses the gated generator by 1.0\% with significantly fewer trainable parameters. 
This indicates that the decoupled weights generated by our $W_D$ and $W_r$ in PYRA are effective and parameter-efficient compared to the common gating strategy.

\section{Conclusion}
\label{sec:5_conclusion}
In this work, we defined and investigated a new challenge named training-inference efficient task adaptation, in which the inference efficiency of large-scale transformers is enhanced during parameter-efficient task adaptation. We propose a novel Parallel Yielding Re-Activation method (PYRA) to effectively cope with the challenge by modulating token features during token merging. Specifically, PYRA generates decoupled parallel yielding modulation weights, and conducts token modulation through re-activation. Extensive experiments show that PYRA introduces negligible performance drops under low compression rate, and bridges the gap of adverse compression between compressed transformers and small-scale models under high compression rate. In real-world applications, our PYRA is highly suitable to the scenario of transferring large-scale vision transformers to downstream tasks where no small-scale model is presented.

\textbf{Limitations.}
We have not yet validated the effectiveness of PYRA in object detection and image segmentation. We plan to extend these tasks in the future. Meanwhile, it is also applicable to attach pruning methods after training the vision transformers with PYRA to eliminate additional parameters attached by PYRA. We leave that topic to future studies.

\section*{Acknowledgements}
This work was supported by National Science and Technology Major 2022ZD0119401, National Natural Science Foundation of China (Nos. 62271281, 61925107, 62021002). It is also sponsored by CAAI-CANN Open Fund, developed on OpenI Community.

%
%
\bibliographystyle{splncs04}
\bibliography{main}

\newpage
\begin{center}
    \noindent{\LARGE \textbf{Supplementary Material}}
\end{center}

\setcounter{section}{0}
\renewcommand\thesection{\Alph{section}}

\begin{table*}[h]
\selectfont
\begin{center}
\caption{The statistics of the VTAB-1k~\cite{zhai2019large} benchmark.}
\label{tab:vtab_stat}
\hspace{-0.44cm}
\begin{tabular}{l|c|l|c|c|l}
\toprule
Dataset & Description &  Classes & Train size & Val size & Test size \\
\midrule
CIFAR-100~\cite{krizhevsky2009learning} & \multirow{7}{*}{Natural} & 100 & \multirow{7}{*}{800/1000} & \multirow{7}{*}{200} & 10000 \\
Caltech101~\cite{fei2006one} &  &  102 &  &  & 6084 \\
DTD~\cite{cimpoi2014describing} &  &  47 &  &  & 1880 \\
Flowers102~\cite{nilsback2008automated} &  &  102 &  &  & 6149 \\
Pets~\cite{parkhi2012cats} &  &  37 &  &  & 3669 \\
SVHN~\cite{netzer2011reading} &  &  10 &  &  & 26032 \\
Sun397~\cite{xiao2010sun} &  &  397 &  &  & 21750 \\
\midrule
Patch Camelyon~\cite{veeling2018rotation} & \multirow{4}{*}{Specialized} &  2 & \multirow{4}{*}{800/1000} & \multirow{4}{*}{200} & 32768 \\
EuroSAT~\cite{helber2019eurosat} &  &  10 &  &  & 5400 \\
Resisc45~\cite{cheng2017remote} &  &  45 &  &  & 6300 \\
Retinopathy~\cite{graham2015kaggle} &  &  5 &  &  & 42670 \\
\midrule
Clevr/count~\cite{johnson2017clevr} & \multirow{8}{*}{Structured} &  8 & \multirow{8}{*}{800/1000} & \multirow{8}{*}{200} & 15000 \\
Clevr/distance~\cite{johnson2017clevr} &  &  6 &  &  & 15000 \\
DMLab~\cite{beattie2016deepmind} &  &  6 &  &  & 22735 \\
KITTI/distance~\cite{geiger2013vision} &  &  4 &  &  & 711 \\
dSprites/location~\cite{matthey2017dsprites} &  &  16 &  &  & 73728 \\
dSprites/orientation~\cite{matthey2017dsprites} &  &  16 &  &  & 73728 \\
SmallNORB/azimuth~\cite{lecun2004learning} &  &  18 &  &  & 12150 \\
SmallNORB/elevation~\cite{lecun2004learning} &  &  9 &  &  & 12150 \\
\bottomrule
\end{tabular}
\end{center}
\end{table*}

\section{Details for the Evaluation Datasets}
\label{sec:dset_info}

We show the detailed statistics of the Visual Task Adaptation Benchmark (VTAB-1k)~\cite{zhai2019large} in \cref{tab:vtab_stat}. Introduced in~\cite{zhai2019large}, the VTAB-1k benchmark contains 19 tasks from diverse domains: (1) \textit{Natural} images that are captured by standard cameras in real-world scenarios; (2) \textit{Specialized} images that are captured by professional equipment, \eg, remote sensing and medical cameras; (3) \textit{Structured} images that are synthesized from simulated environments. VTAB-1k contains a variety of tasks such as object counting and depth estimation apart from the standard image classification. For each task in VTAB-1k, the images are divided into the training set (800 images), the validation set (200 images), and the test set (the original set). All models are fine-tuned on the train+val set, \ie, 1000 images. The validation set is used for hyperparameter tuning.

\begin{table*}[t]
    \centering
    \setlength{\tabcolsep}{2mm}
    \caption{PEFT hyperparameters for each tested backbone.}
    \begin{tabular}{c|cccc}
    \toprule
         & ViT-B/16 & ViT-L/16 & ViT-L/16 (MAE) & DeiT-B/16 \\
         \midrule
        LoRA $h$ & 8 & 12 & 12 & 8 \\
         \bottomrule
    \end{tabular}
    \label{tab:peft_param}
\end{table*}

\begin{table*}[t]
    \centering
    \setlength{\tabcolsep}{2mm}
    \caption{Training hyperparameters for each tested backbone.}
    \begin{tabular}{c|cccc}
    \toprule
         & ViT-B/16 & ViT-L/16 & ViT-L/16 (MAE) & DeiT-B/16 \\
         \midrule
        optimizer       & AdamW & AdamW & AdamW & AdamW \\
        warmup epochs   & 10 & 10 & 10 & 10 \\
        epochs          & 100 & 100 & 100 & 100 \\
        batch size      & 64 & 32 & 32 & 128 \\
        lr (PEFT)       & 1e-3 & 1e-3 & 1e-3 & 1e-3 \\
        wd              & 1e-4 & 1e-4 & 1e-4 & 1e-4 \\
         \bottomrule
    \end{tabular}
    \label{tab:train_param}
\end{table*}

\begin{table*}[t]
    \centering
    \setlength{\tabcolsep}{2mm}
    \caption{The learning rate choices for $W_D$ and $W_r$ in each tested backbone.}
    \begin{tabular}{l|c}
    \toprule
         & lr choices for $W_D$ and $W_r$ \\
         \midrule
        ViT-B/16       & [1e-5, 3e-5, 1e-4, 3e-4, 1e-3, 3e-3] \\
        ViT-L/16       & [1e-5, 3e-5, 1e-4, 3e-4, 1e-3, 3e-3] \\
        ViT-L/16 (MAE) & [1e-6, 3e-6, 1e-5, 3e-5] \\
        DeiT-B/16         & [1e-5, 3e-5, 1e-4, 3e-4, 1e-3, 3e-3] \\
         \bottomrule
    \end{tabular}
    \label{tab:lr_generator}
\end{table*}

\begin{table*}[t]
    \centering
    \setlength{\tabcolsep}{1mm}
    \caption{The merging schedules for the high compression rate benchmark.}
    \begin{tabular}{l|c}
    \toprule
         & $r$ schedule ([$r^1$, $r^2$,$\cdots$, $r^L$]) \\
         \midrule
        ViT-B/16       & [40,34,30,24,18,14,10,8,4,4,3,3] \\
        ViT-L/16       & [20,19,18,17,15,13,13,12,10,9,8,6,6,4,4,4,3,3,2,2,1,1,1,1] \\
        ViT-L/16 (MAE) & [20,19,18,17,15,13,13,12,10,9,8,6,6,4,4,4,3,3,2,2,1,1,1,1] \\
        DeiT-B/16         & [40,34,30,24,18,14,10,8,4,4,3,3] \\
         \bottomrule
    \end{tabular}
    \label{tab:high_rate_sched}
\end{table*}

\section{More Implementation Details}
\label{sec:impl_info}

We present details of implementing PYRA for training-inference efficient task adaptation.

\textbf{PEFT Hyperparameters.}
For training-inference efficient task adaptation, we implement LoRA~\cite{hu2021lora} with hidden layer dimension $h$ as the PEFT module regarding its simplicity and mergeability. We present the PEFT hyperparameters in \cref{tab:peft_param}. Note that $h$ choices in \cref{tab:peft_param} lead to approximately the same percentage of training parameters compared to the backbone for all vision transformers. We use the same $h$ value for each backbone under all competing methods and our PYRA. 

\textbf{Training Hyperparameters.}
For choices of training hyperparameters, we mainly follow~\cite{zhang2022neural,hao2023consolidator} and only conduct minor adjustments on the training batch size for efficient training on our devices. The hyperparameters are listed in \cref{tab:train_param}. To prevent the token modulation from overly altering tokens, thereby leading to training instability, we adopt different learning rates for the modulation weight generators $W_D$ and $W_r$ introduced in PYRA. Specifically, for each tested model, we tune for the optimal learning rate for $W_D$ and $W_r$ on each task within the corresponding range as listed in \cref{tab:lr_generator}.

\textbf{Changing Compression Rate for PYRA.}
To yield compressed smaller-scale models with different throughput values, we change the compression rate of PYRA by changing the $r$ value, \ie, merged token number, for each layer. 

(1) For the low compression rate benchmark, we simply implement a constant merging schedule which simply merges the same number of tokens for each layer. 
Specifically, we merge 16 tokens in each layer for ViT-B/16 and DeiT-B/16, and merge 8 tokens in each layer for ViT-L/16 and ViT-L/16 (MAE). 

(2) For the high compression rate benchmark, we implement a decreasing schedule that consistently outperforms other schedules on the pre-trained dataset~\cite{bolya2022token}.
To achieve the target throughput value demanded by the high compression rate benchmark, we solve an approximate problem and apply minor adjustments to the solutions based on actually tested throughput values.
Formally, for a vision transformer with $L$ layers, total token number $T$ (here we omit the [CLS] token and the distillation token), and $F$ computational FLOPs, we compress it to a smaller-scale model with $f$ FLOPs by merging $r^l$ tokens in layer $l$. We denote the number of remaining tokens in the last layer as $t$ (usually $t<5$). We calculate the approximate solution $\hat{r}^l$ for the merging schedule as:
\begin{equation}
    \hat{r}^l=\lfloor (g(l-1)-g(l))(T-t) \rfloor,\text{where}\ g(x)=(1-\frac{x}{L})^{\frac{F}{f}-1}.
    \label{eq:high_comp_calc}
\end{equation}
After acquiring the approximate solutions $\hat{r}^l$, we apply minor adjustments based on the actually tested throughput values to yield the demanded speedup. The final adjusted $r^l$ yields a smaller-scale compressed model with similar throughput values as the smaller-scale backbone, as listed in Tab. 3 of the article.
The actually applied $r^l$ values for the high compression rate benchmark are listed in \cref{tab:high_rate_sched}. Note that \cref{eq:high_comp_calc} can also yield the adopted constant merging schedule for the low compression rate benchmark. 

For benchmark results in Sec. 4.2 of the article, we adopt the same merging schedules for PYRA and ToMe~\cite{bolya2022token} to guarantee a fair comparison. 

\begin{table}[t]
\fontsize{7.}{8.7}
\selectfont
\setlength{\tabcolsep}{0.8mm}
\centering
\caption{Comparison of different token pruning method choices on the ViT-B/16 with LoRA modules attached as the PEFT choice.}
\label{tab:other_token_methods}
\vspace{-0.1in}
\begin{tabular}{cccc|c}
\toprule
Compression Rate & Method & \# params & Throughput & Average\\
\midrule
\midrule
\multirow{6}{*}{Low} & EViT\cite{liang2022not} & 0.34\% & 732 & 73.09 \\
 & ToFu\cite{kim2024token} & 0.34\% & 748 & 73.31 \\
 & ATS\cite{fayyaz2022adaptive} & 0.34\% & 727 & 71.16 \\
 & LTMP\cite{bonnaerens2023learned} & 0.34\% & 724 & 52.70 \\
 & ToMe\cite{bolya2022token} & 0.34\% & 753 & \underline{74.10} \\
 & PYRA & 0.35\% & 745 & \textbf{74.69} \\
\midrule
\midrule
\multirow{6}{*}{High} & EViT\cite{liang2022not} & 0.34\% & 1200 & 65.79 \\
 & ToFu\cite{kim2024token} & 0.34\% & 1370 & 70.39 \\
 & ATS\cite{fayyaz2022adaptive} & 0.34\% & 1183 & 61.12 \\
 & LTMP\cite{bonnaerens2023learned} & 0.34\% & 1265 & 48.96 \\
 & ToMe\cite{bolya2022token} & 0.34\% & 1381 & \underline{70.43} \\
 & PYRA & 0.35\% & 1365 & \textbf{72.06} \\
\bottomrule
\end{tabular}
\end{table}

\section{Comparing to More Baseline Methods}
Apart from the baseline methods chosen in the main article, we applied several more token pruning methods to demonstrate that our choice of ToMe\cite{bolya2022token} is a strong baseline in the setting of training-inference efficient task adaptation. Specifically, we have applied EViT\cite{liang2022not}, ToFu\cite{kim2024token}, ATS\cite{fayyaz2022adaptive}, and LTMP\cite{bonnaerens2023learned} with LoRA attached as the learnable PEFT module. As shown in \cref{tab:other_token_methods}, all listed methods underperform the strongest baseline, ToMe\cite{bolya2022token}, well demonstrating that analyzing and evaluating ToMe in training-inference efficient task adaptation is important and appropriate. Note that although \cite{kim2024token,bonnaerens2023learned} are follow-up works of ToMe, they involve heavy token pruning that requires extensive training to restore performance, hence underperforming ToMe in PEFT training.

\section{Comparing to Parameter-Efficient Model Distillation}
\label{sec:comp_dist}

\begin{table}[t]
\fontsize{7.}{8.7}
\selectfont
\setlength{\tabcolsep}{0.8mm}
\centering
\caption{Comparison to PE-Dist under the high compression rate.}
\label{tab:exp_dist}
\vspace{-0.1in}
\begin{tabular}{c|ccc|ccc|c}
\toprule
Backbone & Method & \# params & Throughput & Natural & Specialized & Structured & Average\\
\midrule
\midrule
\multirow{2}{*}{ViT-S/16} & PEFT & 0.34\% & 1350 & \textbf{76.29} & \underline{83.56} & \underline{55.71} & \underline{71.85} \\
 & PE-Dist & 0.34\% & 1350 & \textbf{76.29} & \textbf{83.84} & 52.85 & 70.99 \\
\midrule
ViT-B/16 & PYRA & 0.35\% & 1365 & \underline{73.91} & 82.60 & \textbf{59.66} & \textbf{72.06} \\
 \midrule
\midrule
\multirow{2}{*}{ViT-B/16} & PEFT & 0.34\% & 425 & 79.45 & \underline{84.43} & \underline{60.39} & \underline{74.76} \\
 & PE-Dist & 0.34\% & 425 & \underline{80.30} & 84.24 & 59.26 & 74.60 \\
\midrule
ViT-L/16 & PYRA & 0.40\% & 427 & \textbf{80.43} & \textbf{85.17} & \textbf{61.39} & \textbf{75.66} \\
\bottomrule
\end{tabular}
\end{table}

Model distillation~\cite{hinton2015distilling} is an alternative approach for acquiring a smaller-scale model of the desired size. Existing approaches to model distillation~\cite{hinton2015distilling,ahn2019variational,tung2019similarity} require training all parameters in the smaller-scale model with a combined loss:
\begin{equation}
    \mathcal{L}=\mathcal{L}_{\text{task}}+\mathcal{L}_{\text{dist}},
    \label{eq:loss_dist}
\end{equation}
where $\mathcal{L}_{\text{task}}$ represents the task-specific loss, and $\mathcal{L}_{\text{dist}}$ denotes the distillation loss, typically measured by the KL-divergence.

In the setting of training-inference efficient task adaptation, however, training all parameters in the smaller-scale model is unacceptable. Therefore, we implement a baseline as \textit{parameter-efficient model distillation} (PE-Dist), in which we \textit{attach PEFT modules to existing pre-trained smaller-scale backbones, and only train the PEFT modules during model distillation}. We conduct experiments on the ViT backbones~\cite{dosovitskiy2020image}, \ie, ViT-B/16 and ViT-L/16, by distilling them to the smaller-scale backbone, \ie, ViT-S/16 and ViT-B/16, correspondingly. To yield the distillation source, we first train ViT-B/16 and ViT-L/16 by PEFT. As a fair comparison, we choose LoRA~\cite{hu2021lora} as the PEFT method and compare the results for PE-Dist with our PYRA under the high compression rate. As shown in \cref{tab:exp_dist}, parameter-efficient model distillation consistently underperforms directly fine-tuning the smaller-scale model for both backbones, indicating that the \textit{adverse compression} phenomenon still exists when applying model distillation for training-inference efficient task adaptation. In contrast to model distillation, our PYRA consistently eliminates the adverse compression gap and outperforms model distillation by over >1\% on both backbones as well. This indicates that PYRA represents an effective alternative to model distillation for acquiring smaller-scale inference-efficient models under the constraints of parameter-efficient training. 

\section{More Analysis Experiments.}

\begin{table}[t]
\fontsize{7.}{8.7}
\selectfont
\setlength{\tabcolsep}{0.8mm}
\centering
\caption{Comparison of different pipelines on the ViT-B/16.}
\label{tab:pipeline}
\vspace{-0.1in}
\begin{tabular}{ccccc|c}
\toprule
Compression Rate & Pipeline & Method & \# params & Throughput & Average\\
\midrule
\midrule
\multirow{2}{*}{Low} & One-Stage & \multirow{2}{*}{ToMe} & 0.34\% & 753 & \textbf{74.10} \\
 & Two-Stage &  & 0.34\% & 753 & 72.73 \\
\midrule
\midrule
\multirow{2}{*}{High} & One-Stage & \multirow{2}{*}{ToMe} & 0.34\% & 1381 & \textbf{70.43} \\
 & Two-Stage &  & 0.34\% & 1381 & 69.25 \\
\bottomrule
\end{tabular}
\end{table}

\textbf{Is a Two-Stage Schedule Better?}
As discussed in Sec. 3.1 in the article, we employ a one-stage pipeline in which LoRA modules are trained while token merging is attached to the vision transformer. In fact, a more straightforward approach is to implement in a two-stage pipeline: we train the LoRA modules without token merging first, and then employ token merging in the second stage to reduce the redundant tokens. To compare the performance of both pipelines, we apply the ViT-B/16 model pre-trained on ImageNet-21K~\cite{deng2009imagenet} with ToMe attached as the compression method. As shown in \cref{tab:pipeline}, under both low and high compression rates, the one-stage pipeline described in the main article outperforms the two-stage pipeline, well demonstrating that our pipeline choice is appropriate.

\begin{table}[t]
\fontsize{7.}{8.7}
\selectfont
\setlength{\tabcolsep}{0.8mm}
\centering
\makeatletter\def\@captype{table}\makeatother\caption{Adaptation performance comparison between conducting token modulation via PYRA and simply increasing the training parameters in LoRA for the baseline.}.
\label{tab:analysis_param}
\begin{tabular}{cccc|ccc|cc}
\toprule
Method & $h$ & \# params & $\Delta$ params & Natural & Specialized & Structured & Average & $\Delta$ Acc. \\
\midrule
Baseline & 8 & 0.34\% & 1.00$\times$ & \underline{72.9} & 80.8 & 57.6 & 70.43 & 0 \\
Baseline & 16 & 0.69\% & 2.00$\times$ & 72.8 & 80.8 & 58.2 & 70.63 & +0.20 \\
Baseline & 32 & 1.37\% & 4.00$\times$ & \underline{72.9} & \underline{81.9} & \underline{58.3} & \underline{71.00} & +0.57 \\
PYRA & 8 & 0.35\% & 1.03$\times$ & \textbf{73.9} & \textbf{82.6} & \textbf{59.7} & \textbf{72.06} & +1.63 \\
\bottomrule
\end{tabular}
\end{table}

\textbf{Impact of Adding Trainable Parameters.}
To verify that the performance gain of PYRA does not come from the increase of training parameters, 
we compare PYRA with simply increasing the hidden dimension $h$ of LoRA modules in the baseline. 
As shown in \cref{tab:analysis_param}, the baseline method with increased $h$ could only lead to minor performance gains. With 4$\times$ training parameters, the baseline still underperforms PYRA with only $1.03\times$ training parameters by over 1.0\%. 
These results well demonstrate the effectiveness of designs in PYRA. 

\begin{wrapfigure}{R}{0.5\textwidth}
  \vspace{-0.192in}
   \includegraphics[width=0.5\textwidth]{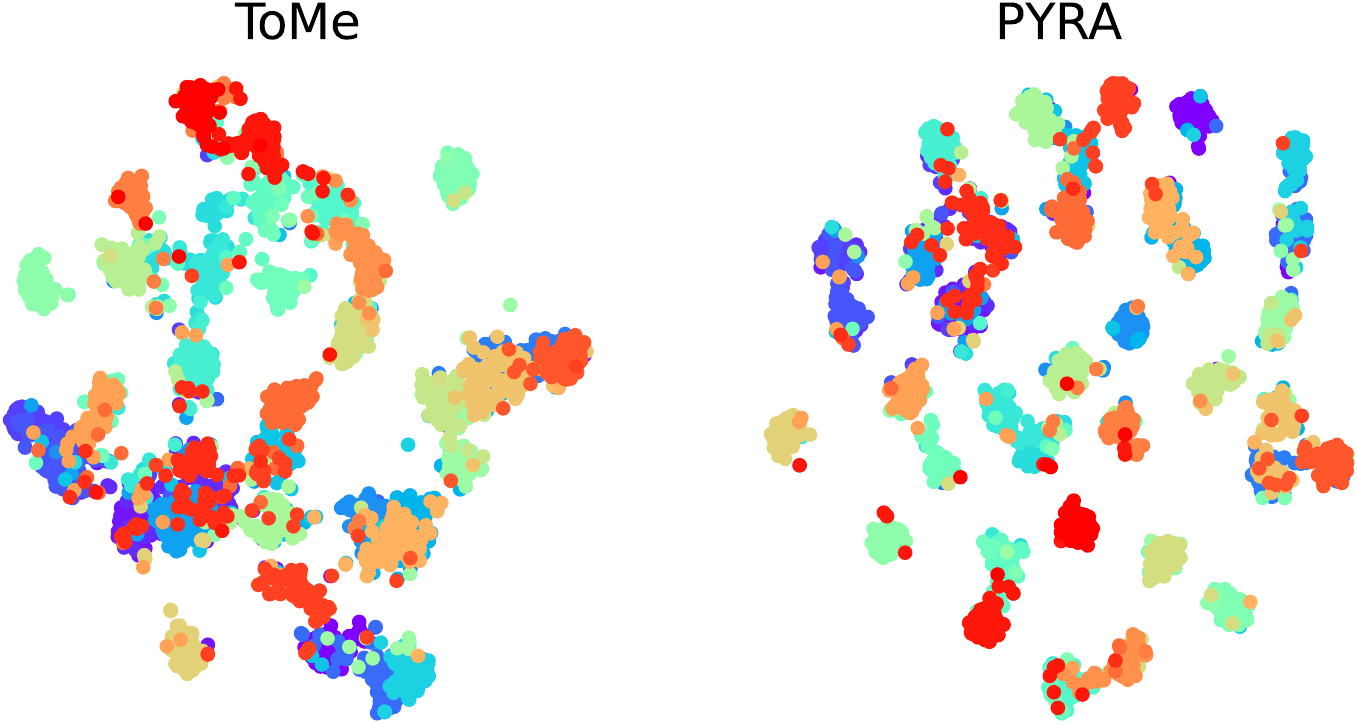}
   \caption{t-SNE~\cite{van2008visualizing} visualization of ToMe~\cite{bolya2022token} and PYRA on the Oxford Pet~\cite{parkhi2012cats} dataset.}
   \label{fig:tsne}
\end{wrapfigure}

\textbf{Qualitative Analysis.}
To qualitatively analyze the effectiveness of our PYRA, we visualize the [CLS] token feature extracted by the strongest ToMe~\cite{bolya2022token} baseline and our PYRA under high compression rate via t-SNE~\cite{van2008visualizing}. As shown in \cref{fig:tsne}, PYRA yields features with clearer clusters compared to ToMe, demonstrating that PYRA can construct a better feature space for downstream tasks.

\begin{figure}
    \centering
    \vspace{-0.1in}
    \includegraphics[width=0.5\textwidth]{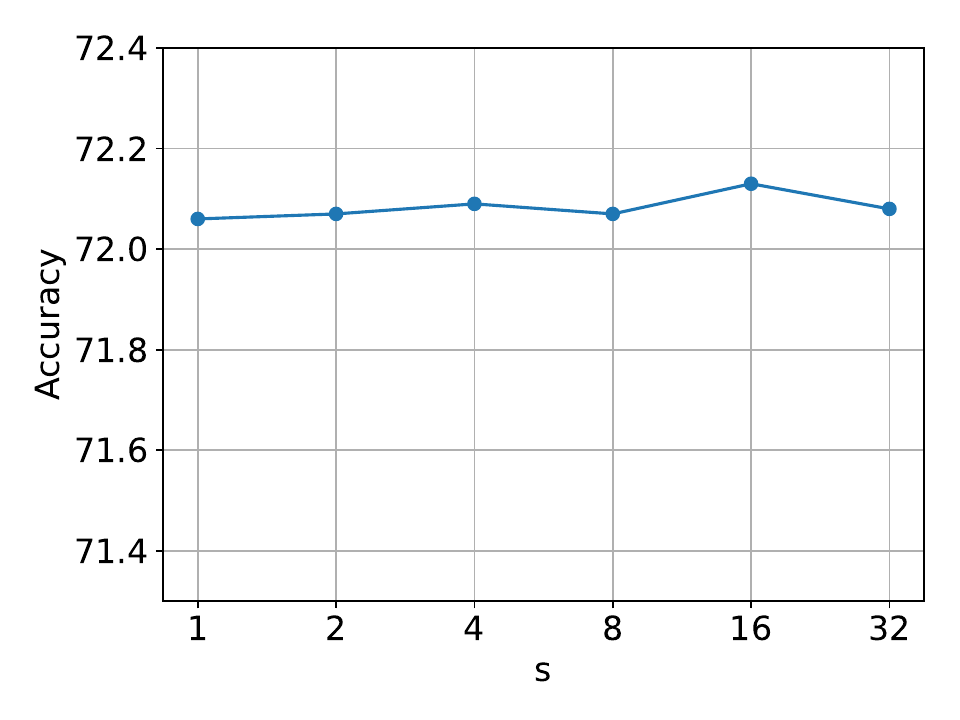}
    \caption{Comparison of different modulation weight generator rank $s$ in PYRA.}
    \vspace{-0.192in}
    \label{fig:analysis_generator_s}
\end{figure}

\textbf{Modulation Matrix Rank.} 
In PYRA, we insert a pair of modulation weight generators $W_r\in \mathbb{R}^{r\times 1}$ and $W_D\in \mathbb{R}^{1\times D}$ for each layer. To prove that setting the matrix rank $s$ of both generators as 1 is enough, we increase the rank of generators and create PYRA variants with more training parameters for comparisons. To apply modulation weights with $s>1$ to the merged tokens, we first calculate $W^l=\delta_D\delta_r$, and then apply token modulation by omitting Eq. 6 of the article and replacing $\hat{\delta}^l_r$ in Eq. 7 of the article to $W^l$. We compare our PYRA ($s=1$) with other variants, \ie, $s=2$, $s=4$, and $s=8$ following Sec. 4.4 in the article, where we apply the ViT-B/16 model pre-trained on ImageNet-21K~\cite{deng2009imagenet} under the high compression rate. As shown in \cref{fig:analysis_generator_s}, although introducing more training parameters, increasing $s$ yields performance comparable to our PYRA with $s=1$. This indicates that simply applying $W_r\in \mathbb{R}^{r\times 1}$ and $W_D\in \mathbb{R}^{1\times D}$ is effective and training-efficient to generate adaptive modulation weights that optimally modulate the tokens to be merged.

\textbf{PYRA on other PEFT methods.} 
For training-inference efficient task adaptation, we have mainly implemented LoRA~\cite{hu2021lora} for its simplicity and mergeability. In conditions where the inference speed is not strictly restrained, PEFT methods that cannot merge into the backbone are also applicable in training-inference efficient task adaptation. To further validate the effectiveness of PYRA on other PEFT methods, we implement Adapter~\cite{houlsby2019parameter} as the PEFT method, which inevitably introduces small computational overhead during inference. We validate our PYRA on the ViT backbones~\cite{dosovitskiy2020image} under the high compression rate. 
As shown in \cref{tab:adapter_vtab_base} and \cref{tab:adapter_vtab_large}, PYRA achieves the best overall performance and outperforms all competing methods in each category. While surpassing the throughput of the smaller-scale backbone with only $\sim$0.2\% training parameters, our PYRA eliminates the adverse compression gap between the compressed larger-scale backbone and the smaller-scale backbone.
This indicates that our PYRA well generalizes to other PEFT methods in terms of achieving training-inference efficient task adaptation.
Experiments on more recent PEFT methods listed in \cref{tab:other_peft_methods} also demonstrate that PYRA consistently outperforms the strongest baseline, ToMe\cite{bolya2022token}, under a broad range of PEFT approaches.

\begin{table}[t]
\begin{minipage}{0.49\linewidth}
\fontsize{7.}{8.7}
\selectfont
\setlength{\tabcolsep}{0.8mm}
\centering
\makeatletter\def\@captype{table}\makeatother\caption{Adaptation results for ViT-Base/16 under high compression rate with the Adapter~\cite{houlsby2019parameter} method. $*$: As a comparison of similar throughputs, we compare ViT-Small/16 with PEFT attached.}
\label{tab:adapter_vtab_base}
\resizebox{\textwidth}{!}{
\begin{tabular}{ccc|ccc|c}
\toprule
\rotatebox{90}{Method} & \rotatebox{90}{\# params} & \rotatebox{90}{Throughput} & \rotatebox{90}{Natural} & \rotatebox{90}{Specialized} & \rotatebox{90}{Structured} & \rotatebox{90}{Average}\\
\midrule
PEFT$*$ & 0.19\% & 1228 & 72.74 & 83.79 & 55.93 & 70.82 \\
\hline
RaP~\cite{DBLP:conf/cvpr/LiA0GTG22} & 0.70\% & 989 & 32.31 & 69.52 & 25.94 & 42.59 \\
DiffRate~\cite{chen2023diffrate} & 0.19\% & 1244 & 48.18 & 70.55 & 23.67 & 47.46 \\
ToMe~\cite{bolya2022token} & 0.18\% & 1301 & \underline{72.37} & \underline{80.57} & \underline{56.06} & \underline{69.66} \\
PYRA & 0.19\% & 1268 & \textbf{74.07} & \textbf{82.25} & \textbf{57.46} & \textbf{71.26} \\
\bottomrule
\end{tabular}}
\end{minipage}\hfill
\begin{minipage} {0.49\linewidth}
\fontsize{7.}{8.7}
\selectfont
\setlength{\tabcolsep}{0.8mm}
\centering
\makeatletter\def\@captype{table}\makeatother\caption{Adaptation results for ViT-Large/16 under high compression rate with the Adapter~\cite{houlsby2019parameter} method. $*$: As a comparison of similar throughputs, we compare ViT-Base/16 with PEFT attached.}
\label{tab:adapter_vtab_large}
\resizebox{\textwidth}{!}{
\begin{tabular}{ccc|ccc|c}
\toprule
\rotatebox{90}{Method} & \rotatebox{90}{\# params} & \rotatebox{90}{Throughput} & \rotatebox{90}{Natural} & \rotatebox{90}{Specialized} & \rotatebox{90}{Structured} & \rotatebox{90}{Average}\\
\midrule
PEFT$*$ & 0.18\% & 393 & 79.33 & 84.69 & 57.41 & 73.81 \\
\hline
RaP~\cite{DBLP:conf/cvpr/LiA0GTG22} & 0.46\% & 240 & 20.24 & 62.89 & 30.14 & 37.76 \\
DiffRate~\cite{chen2023diffrate} & 0.20\% & 391 & 49.60 & 45.79 & 22.47 & 39.29 \\
ToMe~\cite{bolya2022token} & 0.20\% & 410 & \underline{78.84} & \underline{83.63} & \underline{56.90} & \underline{73.12} \\
PYRA & 0.21\% & 402 & \textbf{79.39} & \textbf{84.31} & \textbf{58.00} & \textbf{73.90} \\
\bottomrule
\end{tabular}}
\end{minipage}\hfill
\end{table}

\begin{table}[t]
\fontsize{7.}{8.7}
\selectfont
\setlength{\tabcolsep}{0.8mm}
\centering
\caption{Comparison of different PEFT method choices on the ViT-B/16. Results on ToMe and PYRA are reported.}
\label{tab:other_peft_methods}
\vspace{-0.1in}
\begin{tabular}{cccc}
\toprule
Compression Rate & PEFT choice & ToMe & PYRA \\
\midrule
\midrule
\multirow{3}{*}{Low} & Convpass\cite{jie2022convolutional} & 75.05 & 76.10 \\
 & FacT\cite{jie2023fact} & 74.24 & 75.34 \\
 & SNF\cite{wang2023adapting} & 62.65 & 65.25 \\
\midrule
\midrule
\multirow{3}{*}{High} & Convpass\cite{jie2022convolutional} & 72.09 & 73.26 \\
 & FacT\cite{jie2023fact} & 71.48 & 72.48 \\
 & SNF\cite{wang2023adapting} & 58.01 & 62.49 \\
\bottomrule
\end{tabular}
\end{table}

\section{PYRA Performance on the Pre-Training Task}
\label{sec:other_settings}

PYRA maintains discriminative information with adaptive token modulation, leading to improved performance while achieving complexity reduction via token merging. Extensive experiments have shown the effectiveness of PYRA in the challenge of training-inference efficient task adaptation. Apart from the task adaptation scenario, PYRA is also applicable to the pre-trained model for improving the performance of compressed backbones on the pre-training task.

\begin{table}[t]
\fontsize{7.}{8.7}
\selectfont
\setlength{\tabcolsep}{0.8mm}
\centering
\makeatletter\def\@captype{table}\makeatother\caption{Performance comparison on the ImageNet-1k dataset~\cite{deng2009imagenet}. ``Plain'' denotes directly fine-tuning the MAE pre-trained backbone on ImageNet-1k as in \cite{he2022masked}.}
\label{tab:res_mae_pretrain}
\begin{tabular}{c|cccccc}
\toprule
Model & Method & \# Param & Throughput & Speedup & Acc. & $\Delta$ Acc. \\
\midrule
\multirow{3}{*}{ViT-Base/16 (MAE)} & Plain~\cite{he2022masked} & 86M & 425 & 1.0$\times$ & 83.66 & 0 \\
 & ToMe~\cite{bolya2022token} & 86M & 1381 & 3.2$\times$ & 76.67 & -6.99 \\
 & PYRA & 86M+9.4K & 1365 & 3.2$\times$ & 77.11 & -6.55 \\
\midrule
\multirow{3}{*}{ViT-Large/16 (MAE)} & Plain~\cite{he2022masked} & 303M & 130 & 1.0$\times$ & 85.95 & 0 \\
& ToMe~\cite{bolya2022token} & 303M & 431 & 3.3$\times$ & 81.61 & -4.34 \\
& PYRA & 303M+25K & 427 & 3.3$\times$ & 82.36 & -3.59 \\
\bottomrule
\end{tabular}
\end{table}

We evaluate the effectiveness of PYRA on the pre-training task. Specifically, we follow \cite{bolya2022token} and apply PYRA during the fine-tuning process of self-supervised MAE backbones~\cite{he2022masked}, \ie, ViT-Base/16 (MAE) and ViT-Large/16 (MAE). We evaluate our PYRA under the challenging high compression rate. During fine-tuning, we simply tune the modulation weight generators with other trainable parameters under the same hyperparameters following~\cite{he2022masked}. 
The results are shown in \cref{tab:res_mae_pretrain}. 
Under the high compression rate, ToMe~\cite{bolya2022token} introduces significant performance drops.  With $\sim$0.01\% extra training parameters compared to the backbone, our PYRA significantly compensates the accuracy loss with adaptive token modulation. This can be attributed to that although our PYRA is specially designed to enhance the feature distribution in downstream tasks, it still brings improvements to the pre-training task where full fine-tuning is demanded, which indicates that the insight of improving the perception of feature distribution via token modulation inside our PYRA can be widely adopted to various scenarios. 

\end{document}